\documentclass[acmtog]{acmart}

\AtBeginDocument{%
  }

\usepackage{colortbl, multirow}
\usepackage{xspace}
\newif\ifwithsupp \withsupptrue
\ifwithsupp\usepackage{titletoc}\usepackage{algorithm}\usepackage{algpseudocode}\usepackage{placeins}\fi

\newcommand{\methodName}{X-Splat\xspace}
\definecolor{purple}{rgb}{0.6, 0, 0.6}
\definecolor{darkgreen}{rgb}{0, 0.3, 0}
\definecolor{bestcolor}{HTML}{FFCCCC}
\definecolor{secondcolor}{HTML}{FFE5CC}

\copyrightyear{2026}
\acmYear{2026}
\setcopyright{cc}
\setcctype{by}
\acmConference[SA Conference Papers '26]{SIGGRAPH Asia 2026 Conference Papers}{December 01--04, 2026}{Kuala Lumpur, Malaysia}
\acmBooktitle{SIGGRAPH Asia 2026 Conference Papers (SA Conference Papers '26), December 01--04, 2026, Kuala Lumpur, Malaysia}
\acmDOI{10.1145/3829340.3842177}
\acmISBN{979-8-4007-2842-6/2026/12}

\begin{document}

\title{X-Splat: Unified Gaussian Splatting for Joint RGB and X-ray 3D Reconstruction}

\author{Wangze Xu}
\orcid{0009-0003-7048-3562}
\affiliation{%
 \institution{The University of Tokyo}
 \country{Japan}}
\email{xwz@g.ecc.u-tokyo.ac.jp}

\author{Chu Zhou}
\orcid{0000-0002-3962-6313}
\affiliation{%
 \institution{National Institute of Informatics}
 \country{Japan}}
\email{zhou_chu@pku.edu.cn}

\author{Mae Yamaguchi}
\orcid{0009-0006-8748-3114}
\affiliation{%
 \institution{The University of Tokyo}
 \country{Japan}}

\author{Linrui Dai}
\orcid{0009-0005-1026-6472}
\affiliation{%
 \institution{The University of Tokyo}
 \country{Japan}}
 
\author{I-Chao Shen}
\orcid{0000-0003-4201-3793}
\affiliation{%
 \institution{The University of Tokyo}
 \country{Japan}}
 
\author{Takashi Ijiri}
\orcid{0000-0002-9057-2066}
\affiliation{%
 \institution{Shibaura Institute of Technology}
 \country{Japan}}

\author{Imari Sato}
\orcid{0000-0003-2067-3918}
\affiliation{%
  \institution{National Institute of Informatics}
  \country{Japan}}
\affiliation{%
  \institution{The University of Tokyo}
  \country{Japan}}
\email{imarik@nii.ac.jp}

\begin{abstract}
Jointly reconstructing RGB and X-ray provides new insights by connecting external appearance with internal structure, enabling a more complete understanding of real-world objects than single modality. Yet existing methods train separate models per modality and suffer from redundant geometry, alignment artifacts, and loose coupling that limits cross-modal insight discovery. We propose a unified representation in which all Gaussian primitives share a single set of geometric parameters, with only surface-proximal primitives additionally carrying appearance attributes. We perform cross-modal registration through clean multi-view stereo and X-ray geometric proxies instead of aligning two sets of optimized Gaussians, improving alignment stability. We then jointly optimize the unified representation via supervision from both modalities, complemented by a surface-depth consistency loss derived from X-ray geometry to guide RGB reconstruction. Experiments on the InsideOut dataset and our \textit{Flower \& Insect} dataset show that our framework achieves more robust cross-modal alignment, higher RGB rendering quality without compromising X-ray reconstruction, and a more compact representation than maintaining two independent geometries. Project page: \url{https://zezeaaa.github.io/X-Splat/}.
\end{abstract}



\begin{teaserfigure}
  \includegraphics[width=1.0\textwidth]{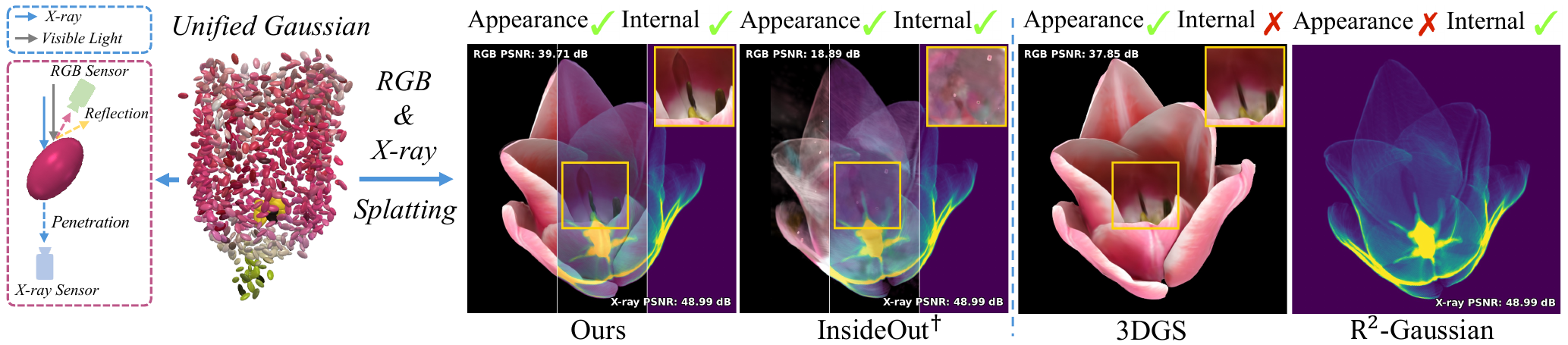}
  \caption{We propose \textit{\methodName{}}, a unified representation where \emph{one} set of Gaussian primitives carries both reflective (RGB) and absorptive (X-ray) attributes, simultaneously capturing external appearance and internal structure while enabling joint optimization. Single-modality specialists (3DGS~\cite{kerbl20233d}, R\textsuperscript{2}-Gaussian~\cite{zha2024r}) cover only one axis; the prior method InsideOut$^\dagger$~\cite{lee2025insideout} covers both modalities, but with lower RGB fidelity in our reproduction (Sec.~\ref{subsec:experimental_setup}). Ours achieves high fidelity on both from a single model. Each Ours/InsideOut panel shows three vertical strips at the same viewpoint: pure RGB / overlay / pure X-ray.
}
  \label{fig:teaser}
\end{teaserfigure}

\maketitle

\section{Introduction}

Real-world objects are both surface and interior: we readily see a flower's outer petals, but not the structure within.  Yet existing 3D reconstruction typically captures only one of the two at a time. RGB cameras record the external appearance through surface reflection, while X-ray imaging penetrates objects to reveal their internal volumetric structure through attenuation, leaving the two as separate digital records of the same physical object. We pursue a unified 3D representation that natively captures both, enabling consistent rendering of an object's exterior and interior from any viewpoint (Fig.~\ref{fig:teaser}). Such a unified representation supports applications such as digital cultural heritage~\cite{bossema2024enabling}, industrial inspection~\cite{bergmann2019mvtec}, and physically-grounded visualization~\cite{ni2024phyrecon}, providing a more complete understanding of real objects than either modality alone.

Existing 3D reconstruction work has developed each modality along a separate track. RGB-based approaches built on neural radiance fields (NeRF)~\cite{mildenhall2021nerf} and Gaussian splatting~\cite{kerbl20233d} reconstruct external surfaces but treat the object interior as unobserved space. X-ray-based methods such as NAF~\cite{zha2022naf}, X-Gaussian~\cite{cai2024radiative}, and R\textsuperscript{2}-Gaussian~\cite{zha2024r} recover internal density distributions but provide no surface appearance. Training separate models per modality is a natural starting point, yet such a decoupled approach keeps two independent representations of the same object with no cross-modal alignment, increasing parameter count and preventing consistent rendering of external appearance and internal structure.

A more principled direction is to design a single framework targeting RGB and X-ray jointly, and the recent InsideOut~\cite{lee2025insideout} is the first such attempt. It trains a separate Gaussian set per modality and bridges them through hierarchical alignment between the two optimized representations. While this design takes a first step toward joint reconstruction, it leaves several aspects underexplored.
First, the two Gaussian sets independently model the same underlying geometry, doubling the parameter footprint.
Second, the alignment operates on optimized Gaussian primitives that are the outcome of geometric and appearance fitting rather than a faithful description of the object's surface, and contain floaters and optimization artifacts, leading to surface-to-interior misalignment that compromises applications requiring aligned internal-external display, such as digital cultural heritage~\cite{bossema2024enabling}.
Finally, although the scene geometry is accurately recovered from X-ray, it is still underutilized in RGB optimization, leaving RGB weakly supervised and limiting surface fidelity in applications such as industrial inspection~\cite{bergmann2019mvtec}.

To address these limitations, we present \textit{\methodName}, a unified Gaussian framework for joint RGB-X-ray 3D reconstruction. Our design is based on a simple observation: although RGB images arise from surface reflection and X-ray images from volumetric attenuation, both are ultimately conditioned on the geometry of the same physical object. This shared geometric origin suggests that the cleanest design is not to model each modality with its own independent geometry, but to let the two modalities share a single geometric representation while specializing only the attributes that genuinely differ between them. Concretely, we propose a unified Gaussian representation in which all primitives share a single set of geometric parameters, and only primitives close to the surface carry appearance attributes, reflecting the physical fact that RGB observations arise primarily from surface reflection. This eliminates the redundancy of maintaining two independent geometries and lets gradients from both modalities flow into the same shared geometry. To bring the two modalities into a common coordinate frame, we register them through clean geometric proxies derived directly from the input data, namely a multi-view stereo (MVS)~\cite{yao2018mvsnet, gu2020cascade, yang2020cost, peng2022rethinking, cao2022mvsformer} point cloud from the RGB views and a mesh extracted from the X-ray reconstruction, rather than aligning the already-optimized Gaussians. Finally, we exploit the shared geometry to enable joint optimization across both modalities, complemented by a surface-depth consistency loss using X-ray-derived surface geometry to supervise RGB rendering. In summary, our contributions are:
\begin{itemize}
    \item \textit{A unified Gaussian representation} with shared geometric
    parameters and surface-restricted appearance attributes,
    enabling compact joint modeling of external appearance and
    internal volumetric structure.
    \item \textit{A geometry-based alignment pipeline} using clean MVS and X-ray proxies to register two modalities, avoiding the surface-to-interior misalignment that arises when registering two sets of already-optimized Gaussians.
    \item \textit{A joint optimization scheme} where both modalities update shared parameters via their respective objectives, complemented by an X-ray-derived surface-depth consistency loss.
\end{itemize}

We evaluate \textit{\methodName} on the InsideOut dataset~\cite{lee2025insideout} and a new \textbf{Flower \& Insect} dataset combining existing flower scans~\cite{ijiri2014flower} with newly captured insect specimens, broadening evaluation to natural objects. Our method achieves more robust cross-modal alignment, better RGB rendering quality without compromising X-ray reconstruction, and is more compact than two independent models.

\section{Related Work}

\paragraph{External Appearance Reconstruction}
\label{subsec:external-recon}

Neural radiance fields (NeRF) \mbox{\cite{mildenhall2021nerf}} have substantially advanced the appearance reconstruction of 3D objects from RGB images, which established a differentiable volume-rendering paradigm that fits a continuous radiance field to multi-view images. Subsequent work improved its rendering quality and efficiency through multi-scale anti-aliasing~\cite{barron2021mip} and hash-grid encodings~\cite{muller2022instant}. A parallel line replaces the volumetric formulation with neural implicit surfaces~\cite{wang2021neus,yariv2021volume} to recover sharper geometry. More recently, 3D Gaussian Splatting~\cite{kerbl20233d} represents the scene with explicit Gaussian primitives and a tile-based rasterizer, achieving real-time rendering at comparable visual quality, with Mip-Splatting~\cite{yu2024mip} addressing aliasing under varying sampling rates. Subsequent work has improved surface fidelity by reformulating the primitives as 2D disks~\cite{huang20242d} or regularizing them toward the underlying surface~\cite{guedon2024sugar}. Other lines extend to sparse-view~\cite{deng2022depth,wang2023sparsenerf,li2024dngaussian,zhu2024fsgs,xu2024mvpgs,peng2024structure}, pose-free~\cite{fan2024instantsplat,fu2024colmap,shen2024disentangled,ye2025no,wang2025vggt}, and dynamic~\cite{fridovich2023k,yang2024deformable,wu20244d,qian20243dgs,hu2024gauhuman,xu2025sequential,zhan2025r3} settings. While these methods achieve high-quality external appearance reconstruction, the internal structure of objects remains beyond their scope.

\paragraph{Internal Structure Reconstruction}
\label{subsec:internal-recon}

X-ray and computed tomography (CT) reconstruction~\cite{gordon1970algebraic, li2023sparse, yu2025x2} recovers the internal density distribution of an object from projection measurements. Classical analytical methods such as FDK~\cite{feldkamp1984practical} remain the standard for cone-beam CT, producing volumetric reconstructions in closed form when projections are dense. To handle sparse-view settings, NAF~\cite{zha2022naf} adapts a neural attenuation field to fit projection measurements with a continuous network representation, and SAX-NeRF~\cite{cai2024structure} employs a line-segment Transformer to capture spatial structure from X-ray projections. More recently, Gaussian splatting has been extended to the X-ray modality. X-Gaussian~\cite{cai2024radiative} replaces alpha-compositing with a radiative rasterizer that follows the Beer-Lambert law, enabling efficient X-ray novel-view synthesis. R\textsuperscript{2}-Gaussian~\cite{zha2024r} further corrects an integration bias in this rasterization formulation and introduces a differentiable voxelizer that enables accurate tomographic reconstruction. Subsequent work also explores direction-disentangled volume rendering for realistic radiographs~\cite{gao2024ddgs}. While these methods achieve high-quality internal density reconstruction, external surface appearance lies beyond the X-ray modality.

\paragraph{Cross-modal 3D Reconstruction}
\label{subsec:cross-modal-recon}

3D reconstruction has long benefited from auxiliary modalities captured by physical sensors that complement RGB cameras. Depth sensors such as LiDAR and commodity time-of-flight devices provide direct geometric measurements that supplement photometric supervision in driving~\cite{rematas2022urban,sun2024lidarf} and indoor scenes~\cite{keetha2024splatam}. Thermal and infrared imaging extend RGB reconstruction to scenarios invisible to RGB cameras, with recent work introducing implicit neural fusion of far-infrared and RGB imagery~\cite{li2024implicit, hassan2024thermonerf}. Multi-spectral and hyperspectral cameras likewise enrich RGB reconstruction with material- and wavelength-specific signals~\cite{thirgood2025hypergs,meyer2025multi}. Recent Gaussian-splatting frameworks further model RGB jointly with thermal images~\cite{lu2024thermalgaussian}, or with multiple modalities through shared and modality-specific Gaussian components~\cite{gu2025mmone}. These auxiliary modalities all carry surface-domain signals, leaving the reconstructed target on the externally observable surface. Few works have attempted to combine external appearance with internal structure. Joining RGB and X-ray for 3D reconstruction differs fundamentally, since they arise from different physical mechanisms. Prior combinations of the two modalities operate at the image level in medical image fusion~\cite{nair2018multi} or serve geometric calibration between the two sensing systems~\cite{albiol2016geometrical}, and learned priors have been used to predict unseen object thickness from RGB-D observations~\cite{nicastro2019x}. None of these produces a unified 3D representation of appearance and internal structure. The most relevant prior work is InsideOut~\cite{lee2025insideout}, which hierarchically fits two independently optimized Gaussian sets and bridges them by a post-hoc visual alignment. In contrast, we align via purely geometric registration and adopt a unified representation with shared geometric parameters, enabling truly joint optimization.

\begin{figure*}[!t]
    \centering
    \includegraphics[width=1.0\linewidth]{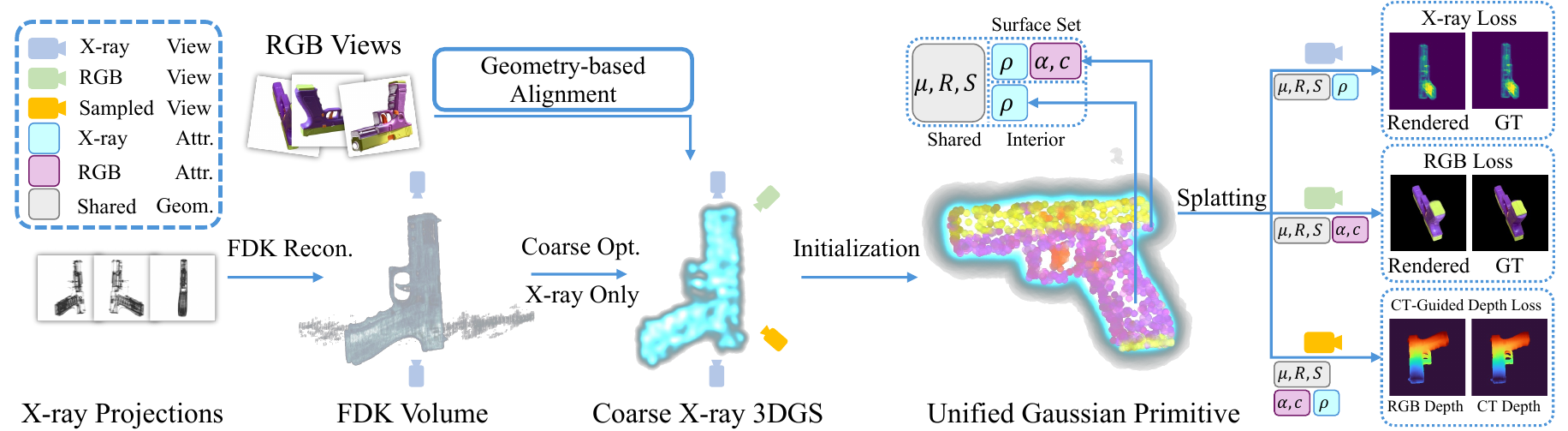}
    \caption{
    \textbf{Framework overview.} A coarse X-ray reconstruction anchors cross-modal alignment (Sec.~\ref{subsec:alignment}) and initializes the unified Gaussian set, in which each primitive carries shared geometry $\{\mu, R, S\}$ and an X-ray attribute $\rho$, with RGB appearance $\{\alpha, c\}$ restricted to a surface subset (Sec.~\ref{subsec:unified-representation}). The unified set is jointly optimized under three losses (right): X-ray photometric, RGB photometric, and a CT-guided depth loss (Sec.~\ref{subsec:joint-optimization}).}
    \label{fig:framework}
\end{figure*}

\section{Preliminaries}
\label{sec:preliminaries}

\paragraph{Imaging Formation: Reflection vs.\ Transmission}
\label{subsec:imaging-formation}

RGB and X-ray imaging differ fundamentally in how light interacts with matter. In RGB imaging, most of the radiance received along a camera ray originates near the first surface it encounters on an opaque object, and the observed pixel intensity is modeled as the integral of emitted color along the ray, weighted by accumulated transmittance,
\begin{equation}
    I_{\mathrm{rgb}}(\mathbf{r}) = \int_{t_{n}}^{t_{f}} T(t)\, \alpha\bigl(\mathbf{r}(t)\bigr)\, c\bigl(\mathbf{r}(t)\bigr)\, dt,
    \:
    T(t) = \exp\!\left(-\int_{t_{n}}^{t} \alpha\bigl(\mathbf{r}(s)\bigr)\, ds\right),
    \label{eq:rgb-volume-rendering}
\end{equation}
where $\alpha(\cdot)$ and $c(\cdot)$ are the volumetric opacity and color fields, $t_{n}$ and $t_{f}$ are the near and far bounds of the ray, and $T(t)$ is the transmittance from the camera to depth $t$. In contrast, an X-ray penetrates the object along its entire path, and the log-domain pixel value is the line integral of the density field $\sigma(\cdot)$ along the ray, governed by the Beer--Lambert law,
\begin{equation}
    I_{\mathrm{xray}}(\mathbf{r}) = \int_{t_{n}}^{t_{f}} \sigma\bigl(\mathbf{r}(t)\bigr)\, dt.
    \label{eq:beer-lambert}
\end{equation}
Despite these distinct optical mechanisms, both modalities observe the same underlying object, and each is a different forward operator acting on a common 3D geometry.

\paragraph{3D Gaussian Representation}
\label{subsec:3d-gaussian}

3D Gaussian splatting~\cite{kerbl20233d} models a scene as $N$ anisotropic 3D Gaussians, each with a center $\mu_{i}$ and a covariance $\Sigma_{i} = R_{i} S_{i} S_{i}^{\top} R_{i}^{\top}$ decomposed into a rotation $R_{i}$ and a diagonal scale $S_{i}$. RGB and X-ray Gaussian splatting share these geometry parameters $\{\mu_{i}, R_{i}, S_{i}\}_{i=1}^{N}$ and differ in their per-primitive optical attributes and rendering operator.

\textit{RGB Gaussian Splatting.} Each primitive additionally carries an opacity $\alpha_{i} \in [0, 1]$ and a view-dependent color $c_{i}$ encoded by spherical harmonics~\cite{kerbl20233d}. Primitives are projected onto the image plane by EWA splatting~\cite{zwicker2001surface}, and the pixel color is obtained by depth-ordered front-to-back compositing of the projected 2D footprints,
\begin{equation}
    I_{\mathrm{rgb}}(\mathbf{r}) = \sum_{i \in \mathcal{N}(\mathbf{r})} c_{i}\, \alpha'_{i} \prod_{j<i}\bigl(1 - \alpha'_{j}\bigr),
    \label{eq:rgb-rendering}
\end{equation}
where $\mathcal{N}(\mathbf{r})$ is the depth-ordered set of primitives intersected by ray~$\mathbf{r}$ and $\alpha'_{i}$ is the EWA-projected effective opacity.

\textit{Radiative Gaussian Splatting.} Each primitive instead carries a single scalar density $\rho_{i} \in \mathbb{R}_{+}$, since X-ray attenuation depends only on isotropic material density. Integrating the resulting density field along the ray according to Eq.~\eqref{eq:beer-lambert} yields, in closed form, a sum of 2D Gaussians on the detector~\cite{cai2024radiative,zha2024r},
\begin{equation}
    I_{\mathrm{xray}}(\mathbf{r}) = \sum_{i=1}^{N} \hat{\rho}_{i}\; \mathcal{N}_{\mathrm{2D}}\!\bigl(\hat{\mathbf{x}};\, \hat{\mu}_{i},\, \hat{\Sigma}_{i}\bigr),
    \label{eq:xray-rendering}
\end{equation}
where $\hat{\mathbf{x}}$, $\hat{\mu}_{i}$, $\hat{\Sigma}_{i}$, and $\hat{\rho}_{i}$ are obtained by projective mapping~\cite{zha2024r}. Here $\mathcal{N}_{\mathrm{2D}}(\cdot;\hat{\mu}_{i},\hat{\Sigma}_{i})$ denotes the 2D Gaussian kernel of the $i$-th projected primitive, and the sum runs over all $N$ primitives.

\section{Method}
\label{sec:method}

Given calibrated RGB views $\{I^{\mathrm{rgb}}_{k}\}_{k=1}^{K_{C}}$ and X-ray projections $\{I^{\mathrm{xray}}_{l}\}_{l=1}^{K_{X}}$ of the same physical object, our goal is to recover a single unified Gaussian set $\{\mathcal{G}_{i}\}_{i=1}^{N}$ that jointly supports high-fidelity RGB novel-view rendering and X-ray novel-projection rendering. As illustrated in Fig.~\ref{fig:framework}, a coarse X-ray reconstruction first anchors cross-modal alignment (Sec.~\ref{subsec:alignment}) and initializes the unified Gaussian set, in which appearance attributes are restricted to a surface subset for compactness (Sec.~\ref{subsec:unified-representation}); the unified set is then jointly optimized under both modalities, bridged by a surface-aware initialization (Secs.~\ref{subsec:joint-optimization}--\ref{subsec:optimization-strategy}).

\subsection{Unified Gaussian Representation}
\label{subsec:unified-representation}

We represent the object by a single set of $N$ Gaussian primitives $\{\mathcal{G}_{i}\}_{i=1}^{N}$. Each primitive $\mathcal{G}_{i}$ carries shared geometry parameters $\{\mu_{i}, R_{i}, S_{i}\}$, a radiative attribute given by a density $\rho_{i} \in \mathbb{R}_{+}$, and, for primitives in a surface subset $\mathcal{S} \subset \{1, \ldots, N\}$, additional appearance attributes consisting of an opacity $\alpha_{i}$ and a view-dependent color $c_{i}$. The shared geometry parameters follow the parametrization in Sec.~\ref{subsec:3d-gaussian} and are used by both renderers, while the radiative and appearance attributes are specific to X-ray and RGB rendering respectively. We write a primitive compactly as
\begin{equation}
    \mathcal{G}_{i} = \bigl\{
    \underbrace{\mu_{i}, R_{i}, S_{i}}_{\text{shared geometry}},\;
    \underbrace{\rho_{i}}_{\text{radiative attribute}},\;
    \underbrace{\{\alpha_{i}, c_{i}\} \text{ if } i \in \mathcal{S}}_{\text{appearance attributes}}
    \bigr\}.
    \label{eq:unified-primitive}
\end{equation}

This restriction has a physical motivation. Restricting appearance attributes to $\mathcal{S}$ reflects the physics of the two modalities. For the opaque objects considered in this work, surface reflection dominates the RGB signal, and interior primitives contribute negligibly to any RGB pixel. Note that only the geometric parameters are shared across modalities; the optical attributes $\rho$ and $\{\alpha, c\}$ remain modality-specific. The construction of $\mathcal{S}$ is detailed in Sec.~\ref{subsec:joint-optimization}.

Sharing the geometry parameters $\{\mu_{i}, R_{i}, S_{i}\}$ across both modalities, rather than maintaining a separate Gaussian set per modality, is motivated by three considerations. \textit{Physically}, RGB reflection and X-ray transmission are different forward operators acting on the same underlying object (Sec.~\ref{subsec:imaging-formation}), so a single shared geometry is more faithful than aligning two independently reconstructed geometries post-hoc. \textit{Compactness-wise}, the shared design allocates $\{\mu, R, S\}$ only once and $\{\alpha, c\}$ only for $i \in \mathcal{S}$, yielding a substantially smaller parameter count than two separate single-modality reconstructions. \textit{Optimization-wise}, gradients from both modalities update the shared geometry, anchoring it under X-ray supervision and compensating for the indirect geometric signal of RGB photometry alone.

\subsection{Geometry-based Alignment}
\label{subsec:alignment}

\begin{figure}[t]
    \centering
    \includegraphics[width=1.0\linewidth]{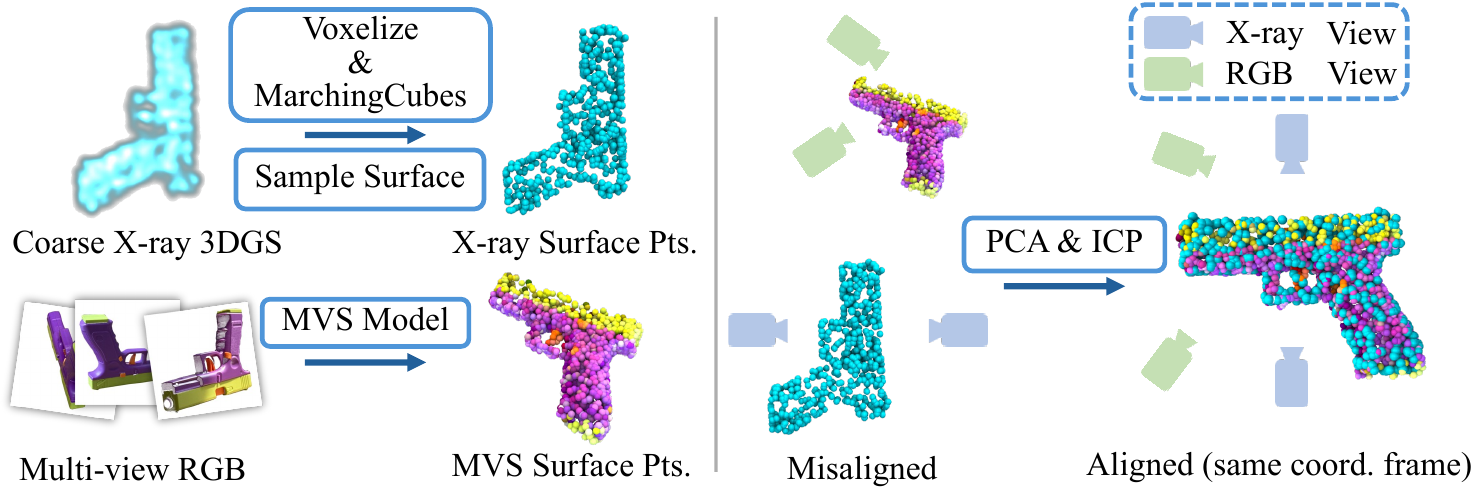}
    \caption{\textbf{Geometry-based alignment.} We extract surface point clouds from each modality independently (left): an X-ray surface $\mathcal{P}^{\mathrm{ct}}$ via Marching Cubes on the coarse X-ray 3DGS, and an MVS surface $\mathcal{P}^{\mathrm{mvs}}$ from multi-view RGB.
    PCA+ICP (right) registers the two surfaces into a shared frame.}
    \label{fig:alignment_pipeline}
\end{figure}

Joint optimization of the unified representation requires the RGB and X-ray modalities to reside in a common coordinate frame. We therefore seek a similarity transformation $T \in \textsc{Sim}(3)$ that maps RGB camera poses into the X-ray coordinate frame. Unlike aligning two sets of optimized Gaussian primitives directly~\cite{lee2025insideout}, which are corrupted by floaters and optimization artifacts, we align on clean geometric proxies extracted from each modality (Fig.~\ref{fig:alignment_pipeline}).

For the RGB side, we run MVSFormer~\cite{cao2022mvsformer} on the calibrated RGB views and their camera parameters $\{I^{\mathrm{rgb}}_{k}, \pi^{\mathrm{rgb}}_{k}\}_{k=1}^{K_{C}}$ to obtain an external surface point cloud $\mathcal{P}^{\mathrm{mvs}} \in \mathbb{R}^{N_{\mathrm{mvs}} \times 3}$. For the X-ray side, we first run a coarse X-ray-only optimization (Sec.~\ref{subsec:joint-optimization}) to obtain a coarse Gaussian set $\mathcal{G}^{\mathrm{coarse}}$, voxelize it via the differentiable voxelizer of R\textsuperscript{2}-Gaussian~\cite{zha2024r} into a CT volume, isolate the outer shell by binarization and morphological closing, extract the CT mesh via Marching Cubes~\cite{lorensen1998marching}, and uniformly sample its surface to obtain the point cloud $\mathcal{P}^{\mathrm{ct}} \in \mathbb{R}^{N_{\mathrm{ct}} \times 3}$.

With these two surface point clouds, we estimate the transformation $T$ by registering $\mathcal{P}^{\mathrm{mvs}}$ against $\mathcal{P}^{\mathrm{ct}}$ in two steps. We first perform principal component analysis (PCA) on each point cloud's positions and align the resulting centroids, principal axes, and variances to obtain an initial similarity transformation $T^{\mathrm{pca}}$ matching translation, rotation, and scale. Starting from $T^{\mathrm{pca}}$, we then apply point-to-point iterative closest point (ICP)~\cite{besl1992method,rusinkiewicz2001efficient} with scaling to obtain the final transformation $T$. We apply $T$ to all RGB camera poses in homogeneous form,
\begin{equation}
    E^{\mathrm{rgb}\to\mathrm{ct}}_{k} = E^{\mathrm{rgb}}_{k}\, T^{-1},
    \label{eq:camera-transform}
\end{equation}
where $E^{\mathrm{rgb}}_{k}$ is the world-to-camera extrinsic of the $k$-th RGB view.

\subsection{Geometry-Constrained Joint Optimization}
\label{subsec:joint-optimization}

The two modalities provide complementary supervision, but X-ray imposes stronger constraints on the shared geometry than RGB. We exploit this asymmetry by establishing the shared geometry from X-ray first, then refining appearance attributes against this anchor.

\textit{Staged schedule.}
A coarse X-ray-only optimization suffices to recover a reliable shared geometry that benefits both the alignment in Sec.~\ref{subsec:alignment} and the subsequent RGB learning. We therefore split the optimization into two stages. In Stage~1, only the X-ray-related parameters $\{\mu_{i}, R_{i}, S_{i}, \rho_{i}\}_{i=1}^{N}$ are optimized under the X-ray photometric loss alone, starting from a standard FDK reconstruction~\cite{feldkamp1984practical,zha2024r}. This produces the coarse X-ray Gaussian set $\mathcal{G}^{\mathrm{coarse}}$ used in Sec.~\ref{subsec:alignment} for CT mesh extraction. In Stage~2, the appearance attributes $\{\alpha_{i}, c_{i}\}_{i \in \mathcal{S}}$ are allocated and initialized as described below, the X-ray-related parameters $\{\mu_{i}, R_{i}, S_{i}, \rho_{i}\}$ are inherited from their Stage~1 values, and all parameters are optimized jointly. We retain the X-ray photometric loss throughout Stage~2 rather than freezing the X-ray side, since RGB gradients could otherwise drift the shared geometry to improve appearance at the cost of X-ray fidelity.

\textit{Surface-aware initialization.}
The surface subset $\mathcal{S}$ introduced in Eq.~\eqref{eq:unified-primitive} partitions the primitives into surface primitives, which carry both the radiative and appearance attributes, and interior primitives, which carry only the shared geometry parameters and the radiative attribute. This partition selectively allocates appearance attributes on top of a shared backbone, rather than splitting the primitives into RGB-only and X-ray-only sets. We construct $\mathcal{S}$ from the CT mesh extracted during alignment to fully leverage the X-ray surface geometry as a prior for RGB optimization, and also to keep the representation compact by preventing interior primitives from participating in RGB rendering. Let $d_{i}$ denote the distance from $\mu_{i}$ to the CT mesh. The surface set is defined by thresholding,
\begin{equation}
    \mathcal{S} = \{\, i : d_{i} < \tau_{d} \,\},
    \label{eq:surface-set}
\end{equation}
where $\tau_{d}$ is a distance threshold in the CT coordinate frame. For $i \in \mathcal{S}$, the opacity is initialized as a Gaussian decay,
\begin{equation}
    \alpha_{i} = \alpha_{\mathrm{surf}} \cdot \exp\!\left(-\frac{d_{i}^{2}}{2\sigma^{2}}\right),
    \label{eq:opacity-init}
\end{equation}
with $\alpha_{\mathrm{surf}}$ a reference opacity and $\sigma$ a decay scale proportional to $\tau_{d}$, so that primitives lying directly on the surface receive the largest initial opacity. The color $c_{i}$ is initialized from the color of the nearest point in $\mathcal{P}^{\mathrm{mvs}}$. The membership of $\mathcal{S}$ is updated throughout training, with details deferred to Sec.~\ref{subsec:optimization-strategy}.

\textit{X-ray-derived geometric constraints.}
The unified representation enables us to render an RGB surface depth $D^{\mathrm{rgb}}$ and a CT reference depth $D^{\mathrm{ct}}$ from a common viewpoint. $D^{\mathrm{rgb}}$ reuses Eq.~\eqref{eq:rgb-rendering} with the along-ray depth $z_i$ in place of $c_i$, restricted to $i \in \mathcal{S}$; $D^{\mathrm{ct}}$ is rasterized from the X-ray mesh, which is periodically re-extracted from the evolving X-ray volume during Stage~2. To exploit the X-ray geometric signal beyond the photometric term without forcing pixel-wise depth agreement, we apply an edge-aware smoothness loss in which the CT depth supplies only edge structure as a stop-gradient target,
\begin{equation}
    \mathcal{L}_{\mathrm{geo}} = \mathbb{E}_{\mathbf{r}}\!\left[ \frac{1}{\bar{D}^{\mathrm{ct}}} \sum_{a\in\{x,y\}} \exp\!\bigl(-|\nabla_{a} D^{\mathrm{ct}}|/\tau\bigr)\,|\nabla_{a} D^{\mathrm{rgb}}| \right],
    \label{eq:geo-loss}
\end{equation}
where $\nabla_{a}$ is the spatial depth gradient along image axis $a$ and $\mathbb{E}_{\mathbf{r}}$ averages over valid foreground pixels, with $\tau$ set to the median of in-mask CT gradient magnitudes and $\bar{D}^{\mathrm{ct}}$ the mean foreground CT depth as a scale normalizer. The loss permits RGB depth discontinuities only where the CT depth has edges, aligning the surface boundary structure of the two depth maps without constraining absolute depth values.

\textit{Overall objective.}
The Stage~1 objective combines the L1 plus D-SSIM X-ray photometric loss $\mathcal{L}^{\mathrm{xray}}_{\mathrm{photo}}$ with the 3D total-variation regularizer $\mathcal{L}_{\mathrm{TV}}$ of R\textsuperscript{2}-Gaussian~\cite{zha2024r},
\begin{equation}
    \mathcal{L}_{\mathrm{stage1}} = \mathcal{L}^{\mathrm{xray}}_{\mathrm{photo}} + \lambda_{\mathrm{tv}} \mathcal{L}_{\mathrm{TV}}.
    \label{eq:stage1-loss}
\end{equation}
The Stage~2 objective adds an L1 plus D-SSIM RGB photometric loss $\mathcal{L}^{\mathrm{rgb}}_{\mathrm{photo}}$ and the geometric consistency loss $\mathcal{L}_{\mathrm{geo}}$,
\begin{equation}
    \mathcal{L}_{\mathrm{stage2}} = \lambda_{\mathrm{rgb}} \mathcal{L}^{\mathrm{rgb}}_{\mathrm{photo}} + \lambda_{\mathrm{xray}} \mathcal{L}^{\mathrm{xray}}_{\mathrm{photo}} + \lambda_{\mathrm{tv}} \mathcal{L}_{\mathrm{TV}} + \lambda_{\mathrm{geo}} \mathcal{L}_{\mathrm{geo}},
    \label{eq:stage2-loss}
\end{equation}
where the $\lambda$'s are scalar loss weights.

\subsection{Adaptive Density Control}
\label{subsec:optimization-strategy}

We extend the adaptive density control of 3DGS~\cite{kerbl20233d} to the unified setting. We maintain two independent position-gradient statistics per primitive, $\bar{g}^{\mathrm{rgb}}_{i}$ and $\bar{g}^{\mathrm{xray}}_{i}$, computed separately for each modality since the two modalities differ in backward frequency and loss scale. In Stage~2, the set of primitives selected for cloning or splitting is
\begin{equation}
    \mathcal{D} = \bigl\{\, \mathcal{G}_{i} \;\big|\; i \in \{1, \ldots, N\}, \;\; \bar{g}^{\mathrm{xray}}_{i} \geq \theta^{\mathrm{xray}} \;\lor\; (\bar{g}^{\mathrm{rgb}}_{i} \geq \theta^{\mathrm{rgb}} \;\land\; i \in \mathcal{S}) \,\bigr\},
    \label{eq:densify-set}
\end{equation}
so that under-fitting in either modality drives the addition of new primitives, where $\theta^{\mathrm{xray}}$ and $\theta^{\mathrm{rgb}}$ are the per-modality gradient densification thresholds. The RGB branch is further restricted to $\mathcal{S}$ since interior primitives carry no appearance attributes; the X-ray branch is unrestricted because X-ray supervision legitimately requires primitives throughout the object interior. New primitives inherit the surface-set membership of their parent. For pruning, removing primitives based on low opacity alone is too aggressive, since a primitive may be invisible to RGB rendering yet carry meaningful X-ray density. We instead require both attributes to be low:
\begin{equation}
    \mathcal{R} = \bigl\{\, \mathcal{G}_{i} \;\big|\; i \in \{1, \ldots, N\}, \;\; \alpha_{i} < \alpha_{\min} \;\land\; \rho_{i} < \rho_{\min} \,\bigr\}.
    \label{eq:prune-set}
\end{equation}
where $\alpha_{\min}$ and $\rho_{\min}$ are the minimum opacity and density thresholds. At each densification interval, primitives in $\mathcal{S}$ whose opacity has fallen below $\alpha_{\min}$ are demoted out of $\mathcal{S}$ rather than pruned, so they no longer participate in RGB rendering but continue contributing to X-ray rendering. In Stage~1, only the X-ray-related criteria apply since no appearance attributes exist yet. The full training procedure is detailed in the supplementary material.
\section{Experiments}
\label{sec:experiments}

\subsection{Experimental Setup}
\label{subsec:experimental_setup}

\paragraph{Datasets.}
We evaluate our method on two datasets covering nine scenes. The first is the InsideOut dataset~\cite{lee2025insideout}, which provides four scenes (Matryoshka, Pharaoh, Skull, and Toy Gun) with paired RGB photographs and ground-truth CT volumes. The second, which we refer to as the \textbf{Flower\,\&\,Insect} dataset, contains two flower scenes (Tulip, Eustoma) from~\cite{ijiri2014flower} and three insect scenes (E02, E15, E32) acquired from prepared specimens, all with paired CT volumes. To evaluate all baselines under identical X-ray acquisition geometry, we follow the protocol of R\textsuperscript{2}-Gaussian~\cite{zha2024r} and re-synthesize cone-beam projections from ground-truth CT volume using TIGRE~\cite{biguri2016tigre}, sampling 50 training and 120 testing projections uniformly over $[0^\circ, 360^\circ)$ at $512\times 512$ resolution. For RGB, we use the real photographs released by InsideOut~\cite{lee2025insideout}, the Blender-rendered views of the real-specimen Flower assets~\cite{ijiri2014flower}, and our captured photographs of the prepared insect specimens, all downsampled to width $800$. We hold out every fourth view for testing and train on the remainder. More details are in the supplementary material.

\paragraph{Metrics.}
For RGB, we report PSNR, SSIM~\cite{wang2004image}, and LPIPS~\cite{zhang2018unreasonable} on held-out test views inside per-view foreground masks. For X-ray, we follow R\textsuperscript{2}-Gaussian~\cite{zha2024r} and report 2D-PSNR/SSIM on rendered projections and 3D-PSNR between the reconstructed and ground-truth CT volumes. For the alignment, we report Chamfer Distance (CD) between the aligned MVS and X-ray surfaces as a direct measure of cross-modal alignment, in normalized scene units where each object's longest extent equals~$2$. For compactness, we report the number of Gaussians (\#G (k) in thousands) and storage (MB).

\paragraph{Implementation Details.}
We implement our framework on top of the R\textsuperscript{2}-Gaussian~\cite{zha2024r} codebase and reuse its X-ray training schedule for Stage~1. Stage~1 runs for $3{,}000$ iterations from an FDK initialization~\cite{feldkamp1984practical}, optimizing $\mathcal{L}^{\mathrm{xray}}_{\mathrm{photo}}$ together with the $3$D total-variation regularizer at $\lambda_{\mathrm{tv}} = 0.05$. After Stage~1 we extract the CT mesh, recover the geometry-based alignment (Sec.~\ref{subsec:alignment}), and construct the surface set with surface distance threshold $\tau_{d} = 0.02$ and reference opacity $\alpha_{\mathrm{surf}} = 0.5$. Stage~2 then jointly optimizes the unified representation for $12{,}000$ iterations under $\lambda_{\mathrm{rgb}} = \lambda_{\mathrm{xray}} = 1.0$, $\lambda_{\mathrm{tv}} = 0.05$, and $\lambda_{\mathrm{geo}} = 0.05$. Each photometric loss combines $\mathcal{L}_1$ with D-SSIM at $\lambda_{\mathrm{dssim}}^{\mathrm{xray}} = 0.25$ and $\lambda_{\mathrm{dssim}}^{\mathrm{rgb}} = 0.2$. The CT mesh is re-extracted every $500$ iterations during Stage~2, and $\mathcal{L}_{\mathrm{geo}}$ is computed on $180$ virtual cameras sampled by Fibonacci sphere. All other hyperparameters follow 3DGS defaults~\cite{kerbl20233d}; complete settings, learning rates, and densification thresholds are listed in the supplementary material. Training takes approximately $14$--$21$ minutes per InsideOut scene on a single NVIDIA RTX A6000 GPU. As the official InsideOut~\cite{lee2025insideout} implementation is not publicly available, we reproduce their method by following the descriptions and hyperparameters reported in their paper. Specifically, InsideOut's final X-ray Gaussians come from its X-ray-only stage 2 (matching R$^2$-Gaussian); we omit its subsequent X-ray refinement because the opacity-based rendering there is incompatible with our density-based renderer. We denote this reproduction by InsideOut$^\dagger$ in all tables and figures. The full reproduction setup and a discussion of discrepancies with its originally reported results are provided in the supplementary material.

\subsection{Per-Modality Reconstruction Quality}
\begin{figure*}[h]
\centering
\includegraphics[width=1.0\textwidth]{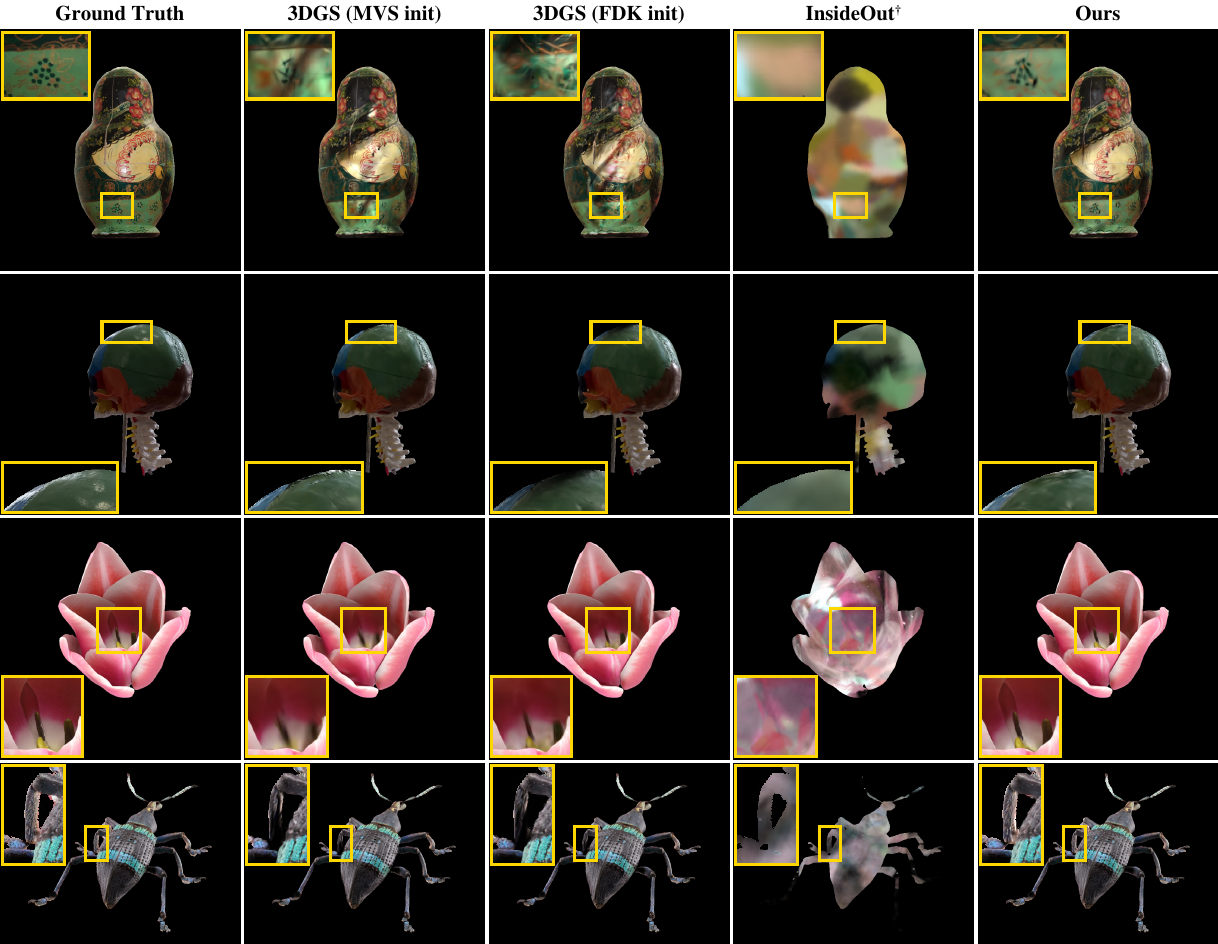}
\caption{\textbf{Qualitative comparison of RGB novel-view synthesis.} Held-out test views across both datasets; yellow boxes mark zoom-in regions.}
\label{fig:qual_rgb}
\end{figure*}

\begin{figure*}[h]
\centering
\includegraphics[width=1.0\textwidth]{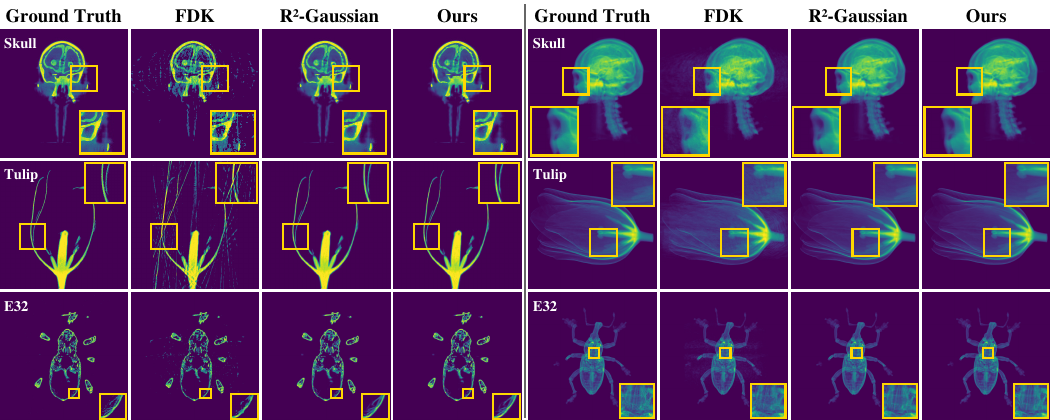}
\caption{\textbf{CT volume slice and X-ray projection comparison.} Left 4 columns: CT slices; right 4: X-ray projections. InsideOut is not shown separately, as its X-ray stage follows R\textsuperscript{2}-Gaussian and coincides with that column.
}
\label{fig:qual_ct_xray}
\end{figure*}

\begin{table*}[h]
\centering
\small
\caption{\textbf{Per-modality reconstruction quality on InsideOut Dataset.} Best in \textbf{bold}, second in \underline{underline}.}
\label{tab:permod_insideout}
\resizebox{1.0\linewidth}{!}{%
\begin{tabular}{l|ccc|ccc|ccc|ccc|ccc}
\toprule
Subject: & \multicolumn{3}{c|}{Matryoshka} & \multicolumn{3}{c|}{Pharaoh} & \multicolumn{3}{c|}{Skull} & \multicolumn{3}{c|}{Toy Gun} & \multicolumn{3}{c}{Average} \\
\midrule
\multicolumn{16}{l}{\textit{(a) RGB Novel View Synthesis}} \\
Metric: & PSNR$\uparrow$ & SSIM$\uparrow$ & LPIPS$\downarrow$ & PSNR$\uparrow$ & SSIM$\uparrow$ & LPIPS$\downarrow$ & PSNR$\uparrow$ & SSIM$\uparrow$ & LPIPS$\downarrow$ & PSNR$\uparrow$ & SSIM$\uparrow$ & LPIPS$\downarrow$ & PSNR$\uparrow$ & SSIM$\uparrow$ & LPIPS$\downarrow$ \\
\midrule
3DGS (MVS init)~\cite{kerbl20233d}           & 29.56 & \underline{0.9643} & \underline{0.0393} & \underline{29.64} & \underline{0.9719} & \underline{0.0291} & \underline{31.08} & \underline{0.9670} & \underline{0.0475} & 34.54 & \underline{0.9871} & \underline{0.0133} & \underline{31.20} & \underline{0.9726} & \underline{0.0323} \\
3DGS (FDK init)~\cite{kerbl20233d}           & \underline{29.75} & 0.9628 & 0.0408 & 29.51 & 0.9707 & 0.0298 & 30.97 & 0.9660 & 0.0501 & \underline{34.55} & 0.9870 & 0.0136 & 31.20 & 0.9716 & 0.0336 \\
InsideOut$^\dagger$~\cite{lee2025insideout}                 & 18.13 & 0.8836 & 0.1522 & 21.78 & 0.9039 & 0.0927 & 19.57 & 0.9003 & 0.1497 & 25.39 & 0.9609 & 0.0481 & 21.22 & 0.9122 & 0.1107 \\
\textbf{Ours}             & \textbf{30.37} & \textbf{0.9693} & \textbf{0.0354} & \textbf{30.89} & \textbf{0.9756} & \textbf{0.0272} & \textbf{32.42} & \textbf{0.9713} & \textbf{0.0433} & \textbf{36.68} & \textbf{0.9900} & \textbf{0.0114} & \textbf{32.59} & \textbf{0.9765} & \textbf{0.0293} \\
\midrule
\multicolumn{16}{l}{\textit{(b) X-ray Reconstruction}} \\
Metric: & 2D-PSNR$\uparrow$ & 2D-SSIM$\uparrow$ & 3D-PSNR$\uparrow$ & 2D-PSNR$\uparrow$ & 2D-SSIM$\uparrow$ & 3D-PSNR$\uparrow$ & 2D-PSNR$\uparrow$ & 2D-SSIM$\uparrow$ & 3D-PSNR$\uparrow$ & 2D-PSNR$\uparrow$ & 2D-SSIM$\uparrow$ & 3D-PSNR$\uparrow$ & 2D-PSNR$\uparrow$ & 2D-SSIM$\uparrow$ & 3D-PSNR$\uparrow$ \\
\midrule
FDK~\cite{feldkamp1984practical}                       & 31.93 & 0.9019 & 24.47 & 42.92 & 0.9829 & 35.65 & 31.84 & 0.7997 & 25.91 & 28.77 & 0.9317 & 28.25 & 33.86 & 0.9040 & 28.57 \\
R$^2$-Gaussian~\cite{zha2024r}            & \underline{48.20} & \underline{0.9940} & \underline{33.35} & \underline{52.33} & \textbf{0.9984} & \textbf{42.28} & \underline{49.56} & \textbf{0.9963} & \textbf{37.09} & \textbf{48.96} & \textbf{0.9966} & \textbf{40.28} & \underline{49.76} & \textbf{0.9963} & \textbf{38.25} \\
InsideOut$^\dagger$~\cite{lee2025insideout}                 & \underline{48.20} & \underline{0.9940} & \underline{33.35} & \underline{52.33} & \textbf{0.9984} & \textbf{42.28} & \underline{49.56} & \textbf{0.9963} & \textbf{37.09} & \textbf{48.96} & \textbf{0.9966} & \textbf{40.28} & \underline{49.76} & \textbf{0.9963} & \textbf{38.25} \\
\textbf{Ours}             & \textbf{48.34} & \textbf{0.9941} & \textbf{33.39} & \textbf{52.51} & \underline{0.9983} & \underline{42.13} & \textbf{49.72} & \textbf{0.9963} & \underline{36.90} & \underline{48.69} & \underline{0.9964} & \underline{40.25} & \textbf{49.81} & \textbf{0.9963} & \underline{38.16} \\
\bottomrule
\end{tabular}%
}
\end{table*}
\begin{table*}[h]
\centering
\small
\caption{\textbf{Per-modality reconstruction quality on Flower \& Insect Dataset.}}
\label{tab:permod_flower_insect}
\resizebox{1.0\linewidth}{!}{%
\begin{tabular}{l|ccc|ccc|ccc|ccc|ccc|ccc}
\toprule
Subject: & \multicolumn{3}{c|}{Tulip} & \multicolumn{3}{c|}{Eustoma} & \multicolumn{3}{c|}{E02} & \multicolumn{3}{c|}{E15} & \multicolumn{3}{c|}{E32} & \multicolumn{3}{c}{Average} \\
\midrule
\multicolumn{19}{l}{\textit{(a) RGB Novel View Synthesis}} \\
Metric: & PSNR$\uparrow$ & SSIM$\uparrow$ & LPIPS$\downarrow$ & PSNR$\uparrow$ & SSIM$\uparrow$ & LPIPS$\downarrow$ & PSNR$\uparrow$ & SSIM$\uparrow$ & LPIPS$\downarrow$ & PSNR$\uparrow$ & SSIM$\uparrow$ & LPIPS$\downarrow$ & PSNR$\uparrow$ & SSIM$\uparrow$ & LPIPS$\downarrow$ & PSNR$\uparrow$ & SSIM$\uparrow$ & LPIPS$\downarrow$ \\
\midrule
3DGS (MVS init)~\cite{kerbl20233d}           & \underline{38.34} & \underline{0.9888} & \underline{0.0322} & \underline{37.98} & \underline{0.9883} & \underline{0.0414} & 30.63 & \underline{0.9743} & \underline{0.0283} & 30.10 & \underline{0.9749} & \underline{0.0246} & \underline{31.55} & \underline{0.9793} & \textbf{0.0234} & \underline{33.72} & \underline{0.9811} & \underline{0.0300} \\
3DGS (FDK init)~\cite{kerbl20233d}           & 37.85 & 0.9874 & 0.0387 & 37.74 & 0.9875 & 0.0466 & \underline{30.88} & 0.9741 & 0.0289 & \underline{30.11} & 0.9742 & 0.0255 & 31.53 & 0.9788 & \underline{0.0239} & 33.62 & 0.9804 & 0.0327 \\
InsideOut$^\dagger$~\cite{lee2025insideout}                 & 18.89 & 0.9148 & 0.1262 & 17.79 & 0.9378 & 0.1184 & 19.82 & 0.8809 & 0.1435 & 19.90 & 0.8748 & 0.1347 & 19.09 & 0.8534 & 0.1611 & 19.10 & 0.8923 & 0.1368 \\
\textbf{Ours}             & \textbf{39.71} & \textbf{0.9917} & \textbf{0.0219} & \textbf{39.22} & \textbf{0.9901} & \textbf{0.0370} & \textbf{34.75} & \textbf{0.9839} & \textbf{0.0246} & \textbf{33.74} & \textbf{0.9844} & \textbf{0.0223} & \textbf{34.77} & \textbf{0.9851} & 0.0253 & \textbf{36.44} & \textbf{0.9871} & \textbf{0.0262} \\
\midrule
\multicolumn{19}{l}{\textit{(b) X-ray Reconstruction}} \\
Metric: & 2D-PSNR$\uparrow$ & 2D-SSIM$\uparrow$ & 3D-PSNR$\uparrow$ & 2D-PSNR$\uparrow$ & 2D-SSIM$\uparrow$ & 3D-PSNR$\uparrow$ & 2D-PSNR$\uparrow$ & 2D-SSIM$\uparrow$ & 3D-PSNR$\uparrow$ & 2D-PSNR$\uparrow$ & 2D-SSIM$\uparrow$ & 3D-PSNR$\uparrow$ & 2D-PSNR$\uparrow$ & 2D-SSIM$\uparrow$ & 3D-PSNR$\uparrow$ & 2D-PSNR$\uparrow$ & 2D-SSIM$\uparrow$ & 3D-PSNR$\uparrow$ \\
\midrule
FDK~\cite{feldkamp1984practical}                       & 30.32 & 0.8487 & 21.94 & 30.61 & 0.8542 & 27.62 & 35.92 & 0.8996 & 28.89 & 38.74 & 0.9514 & 31.13 & 36.54 & 0.9146 & 28.70 & 34.42 & 0.8937 & 27.66 \\
R$^2$-Gaussian~\cite{zha2024r}            & \underline{48.99} & \textbf{0.9958} & \textbf{31.75} & \textbf{53.96} & \textbf{0.9976} & \textbf{40.30} & \textbf{47.89} & \textbf{0.9978} & \underline{35.09} & \textbf{47.07} & \textbf{0.9973} & \textbf{35.28} & \underline{46.98} & \textbf{0.9975} & \textbf{34.13} & \textbf{48.98} & \textbf{0.9972} & \textbf{35.31} \\
InsideOut$^\dagger$~\cite{lee2025insideout}                 & \underline{48.99} & \textbf{0.9958} & \textbf{31.75} & \textbf{53.96} & \textbf{0.9976} & \textbf{40.30} & \textbf{47.89} & \textbf{0.9978} & \underline{35.09} & \textbf{47.07} & \textbf{0.9973} & \textbf{35.28} & \underline{46.98} & \textbf{0.9975} & \textbf{34.13} & \textbf{48.98} & \textbf{0.9972} & \textbf{35.31} \\
\textbf{Ours}             & \textbf{48.99} & \underline{0.9957} & \underline{31.74} & \underline{53.74} & \underline{0.9974} & \underline{40.15} & \underline{47.86} & \underline{0.9978} & \textbf{35.11} & \underline{46.97} & \textbf{0.9973} & \underline{35.24} & \textbf{47.05} & \textbf{0.9975} & \underline{34.12} & \underline{48.92} & \underline{0.9971} & \underline{35.27} \\
\bottomrule
\end{tabular}%
}
\end{table*}

Tables~\ref{tab:permod_insideout} and~\ref{tab:permod_flower_insect} report per-modality novel-view metrics on both datasets, with sub-tables (a) for RGB rendering and (b) for X-ray reconstruction. On the RGB side, Ours achieves consistently competitive metrics across all scenes on both datasets, improving over either 3DGS variant by around 1\,dB PSNR on InsideOut and over 2\,dB on Flower \& Insect. Even FDK initialization, which already encodes the full volumetric structure, fails to outperform the MVS-init baseline, confirming that better initial geometry alone cannot break the single-modality ceiling without persistent geometric supervision. InsideOut$^\dagger$~\cite{lee2025insideout} yields substantially lower RGB quality in our experiments (reproduction setup in Sec.~\ref{subsec:experimental_setup}). On the X-ray side, our representation produces reconstruction quality consistent with R\textsuperscript{2}-Gaussian~\cite{zha2024r} across both datasets, with only minor fluctuations across individual metrics, while the classical FDK reconstruction degrades substantially compared to the Gaussian-based methods. The qualitative results shown in Fig.~\ref{fig:qual_rgb} and Fig.~\ref{fig:qual_ct_xray} suggest that our method recovers sharper surface texture and fewer floater artifacts than all RGB baselines on scenes such as Matryoshka and Tulip, and our reconstructed CT slices and X-ray projections produce details closely matching R\textsuperscript{2}-Gaussian on scenes from both datasets. Joint training over a shared geometry therefore substantially improves RGB reconstruction quality without compromising X-ray reconstruction fidelity, validating the staged optimization of Sec.~\ref{subsec:joint-optimization}.

\subsection{Joint RGB-X-ray Reconstruction}
\begin{table}[h]
\centering
\caption{\textbf{Cross-modal alignment quality on InsideOut Dataset.} The \emph{Input Source} column indicates whether each method operates on optimized Gaussian primitives or on clean geometric proxies (MVS surface + CT mesh).}
\label{tab:align_main}
\resizebox{1.0\linewidth}{!}{%
\begin{tabular}{l|l|c}
\toprule
Method & Input Source & CD $\downarrow$ \\
\midrule
\addlinespace[1pt]
InsideOut$^\dagger$ Hierarchical Alignment~\cite{lee2025insideout} & Optimized Gaussian primitives & 0.0978 \\
\midrule
\addlinespace[1pt]
ICP Only & \multirow{3}{*}{Geometry-based (MVS + CT mesh)} & 0.1453 \\
PCA Only & & 0.0250 \\
\textbf{Ours} (PCA + ICP) & & \textbf{0.0122} \\
\bottomrule
\end{tabular}%
}
\end{table}

\begin{figure*}[h]
\centering
\includegraphics[width=\textwidth]{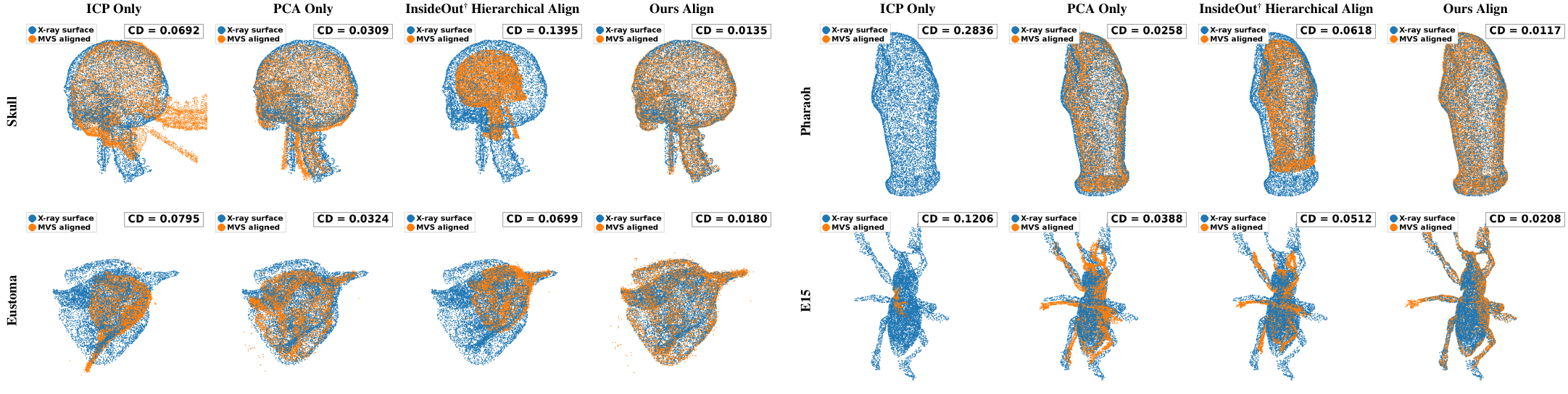}
\caption{\textbf{Cross-modal alignment comparison across both datasets.} Aligned MVS surface (orange) overlaid on X-ray surface (blue); lower CD is better.}
\label{fig:qual_alignment}
\end{figure*}

\textit{Cross-Modal Alignment.}
Table~\ref{tab:align_main} reports Chamfer Distance to the X-ray-reconstructed surface for our alignment pipeline and three representative alternatives, all applied to the same source point cloud and the same target geometry so that only the estimated similarity transformation $T$ varies. As shown by both Table~\ref{tab:align_main} and Fig.~\ref{fig:qual_alignment} (e.g., the residual misalignment on Skull), the InsideOut hierarchical alignment is less accurate than the other three methods. It operates on optimized Gaussian primitives whose floaters and overfit details propagate directly into the alignment. In contrast, aligning the clean MVS surface and CT mesh proxies with PCA alone already outperforms the InsideOut hierarchical alignment, and our full PCA+ICP pipeline improves Chamfer Distance by roughly $8\times$. The geometric cleanliness of the alignment source is therefore more decisive than the alignment algorithm itself. PCA provides global orientation and scale alignment, while ICP refines the result locally. PCA alone is insufficient for fine alignment, and identity-initialized ICP without a good PCA seed can fall into local minima and even collapse to near-degenerate scales on symmetric objects (e.g., E15 in Fig.~\ref{fig:qual_alignment}). The two stages are therefore complementary. 

\begin{figure*}[h]
\centering
\includegraphics[width=\textwidth]{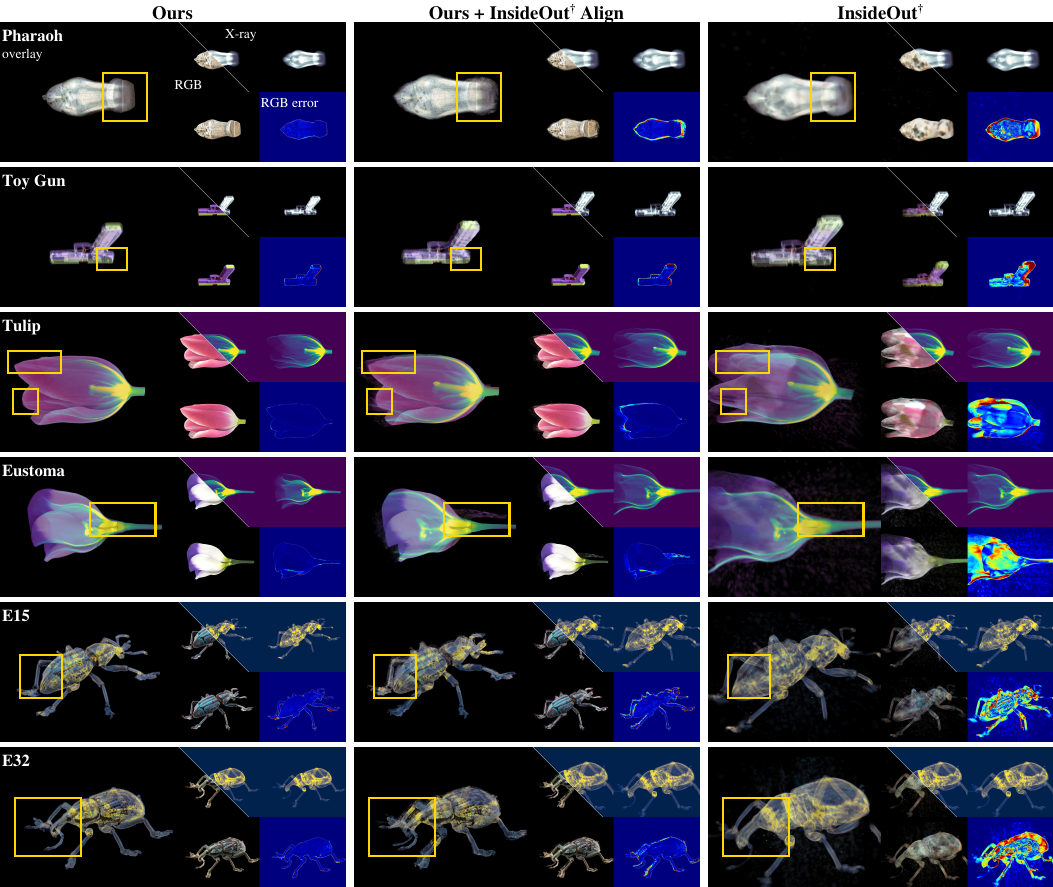}
\caption{\textbf{Unified RGB-X-ray rendering at the same viewpoint.} Each method's cell shows overlay, RGB, X-ray, and an RGB error map at one test view.}
\label{fig:qual_overlay}
\end{figure*}

\textit{Joint Quality and Coherence.}
Table~\ref{tab:joint_summary} aggregates RGB, X-ray, and compactness metrics for our unified representation, InsideOut, and an Independent baseline that trains our X-ray-only and RGB-only variants as two separate models with parameter counts summed. Ours outperforms Independent on RGB by a consistent margin and significantly surpasses InsideOut, while matching Independent on X-ray reconstruction quality. Ours also uses around 26\% less storage than Independent. These savings come from sharing the geometric backbone and restricting appearance attributes to the surface set, while the RGB gain comes from joint X-ray supervision providing geometric constraints that RGB photometry alone cannot offer. Fig.~\ref{fig:qual_overlay} visualizes this by rendering RGB and X-ray from a single model at the same test RGB viewpoint, a setting inherently inaccessible to two independent models. On scenes such as Pharaoh, Tulip, and E15, our method produces sharp surface texture together with clear internal structure in one consistent rendering. The middle column, in which our alignment is replaced by the InsideOut hierarchical alignment (Ours~+~InsideOut~Align), shows visible surface misalignment artifacts consistent with row (d) of Table~\ref{tab:abl_main}, while the InsideOut column exhibits the RGB degradation already evident in the per-modality comparison. Consistently rendering the internal X-ray volume and the external RGB surface from a single model is the key advantage of our unified Gaussian representation. The supplementary video showcases these dynamic renderings.

\begin{table}[h]
\centering
\small
\caption{\textbf{Joint inside-outside reconstruction summary on InsideOut Dataset.} The Independent baseline's RGB-only model uses FDK initialization to match Ours's init source, and its two model sizes are summed.}
\label{tab:joint_summary}
\resizebox{1.0\linewidth}{!}{%
\begin{tabular}{l|ccc|cc|cc}
\toprule
& \multicolumn{3}{c|}{RGB Novel View} & \multicolumn{2}{c|}{X-ray} & \multicolumn{2}{c}{Compactness} \\
Method          & PSNR$\uparrow$ & SSIM$\uparrow$ & LPIPS$\downarrow$ & 2D-PSNR$\uparrow$ & 3D-PSNR$\uparrow$ & \#G (k) & Size (MB) \\
\midrule
Independent (X-ray / RGB only)   & 31.20 & 0.9716 & 0.0336 & 49.76 & 38.25 & 168.8 & 17.28 \\
InsideOut$^\dagger$~\cite{lee2025insideout}     & 21.22 & 0.9122 & 0.1107 & 49.76 & 38.25 & 166.3 & 18.75 \\
\textbf{Ours} (unified)               & \textbf{32.59} & \textbf{0.9765} & \textbf{0.0293} & \textbf{49.81} & \underline{38.16} & \textbf{129.1} & \textbf{12.79} \\
\bottomrule
\end{tabular}%
}
\end{table}

\subsection{Ablation Studies}
\label{sec:ablation}

\begin{table}[h]
\centering
\small
\caption{\textbf{Main ablation study on InsideOut Dataset.} \emph{Components}: \textbf{CT depth}, CT-depth consistency loss (Eq.~\ref{eq:geo-loss}); \textbf{Surf. set}, surface-set mechanism (Eq.~\ref{eq:surface-set}); \textbf{Ours align.}, geometry-based alignment (Sec.~\ref{subsec:alignment}); \textbf{Shared geom.}, unified Gaussian representation (Sec.~\ref{subsec:unified-representation}).}
\label{tab:abl_main}
\resizebox{1.0\linewidth}{!}{%
\begin{tabular}{l|cccc|ccc|cc|cc}
\toprule
\multirow{2}{*}{Configuration}              & \multicolumn{4}{c|}{Components} & \multicolumn{3}{c|}{RGB Novel View} & \multicolumn{2}{c|}{X-ray} & \multicolumn{2}{c}{Compactness} \\
& CT depth & Surf. set & Ours align. & Shared geom. & PSNR$\uparrow$ & SSIM$\uparrow$ & LPIPS$\downarrow$ & 2D-PSNR$\uparrow$ & 3D-PSNR$\uparrow$ & \#G (k) & Size (MB) \\
\midrule
(a). \textbf{Ours} (full joint)              & \checkmark & \checkmark & \checkmark & \checkmark & \textbf{32.59} & \textbf{0.9765} & \textbf{0.0293} & \textbf{49.81} & 38.16 & \textbf{129.1} & \underline{12.79} \\
(b). w/o CT depth                             &            & \checkmark & \checkmark & \checkmark & \underline{32.43} & \underline{0.9757} & 0.0299 & 49.75 & \underline{38.19} & \textbf{129.1} & \textbf{12.24} \\
(c). w/o surface set                          &            &            & \checkmark & \checkmark & 32.41 & 0.9751 & \underline{0.0297} & 49.65 & 38.12 & \underline{129.9} & 31.19 \\
(d). w/o Ours align.                          &            &            &            & \checkmark & 31.50 & 0.9677 & 0.0382 & 49.65 & 38.16 & 131.6 & 30.12 \\
(e). Independent                              &            &            &            &            & 31.20 & 0.9716 & 0.0336 & \underline{49.76} & \textbf{38.25} & 168.8 & 17.28 \\
\bottomrule
\end{tabular}%
}
\end{table}

Table~\ref{tab:abl_main} reports a cumulative ablation on the four InsideOut scenes. The four \emph{Components} columns indicate which design elements are enabled in each configuration: rows (a)--(d) progressively disable individual components from left to right, while row (e) trains the single-modality variants of our framework as two separate models. CT-depth supervision and the surface-set mechanism both rely on an accurate cross-modal alignment, so they remain disabled once alignment is replaced in row (d).

\begin{figure}[h]
\centering
\includegraphics[width=1.0\linewidth]{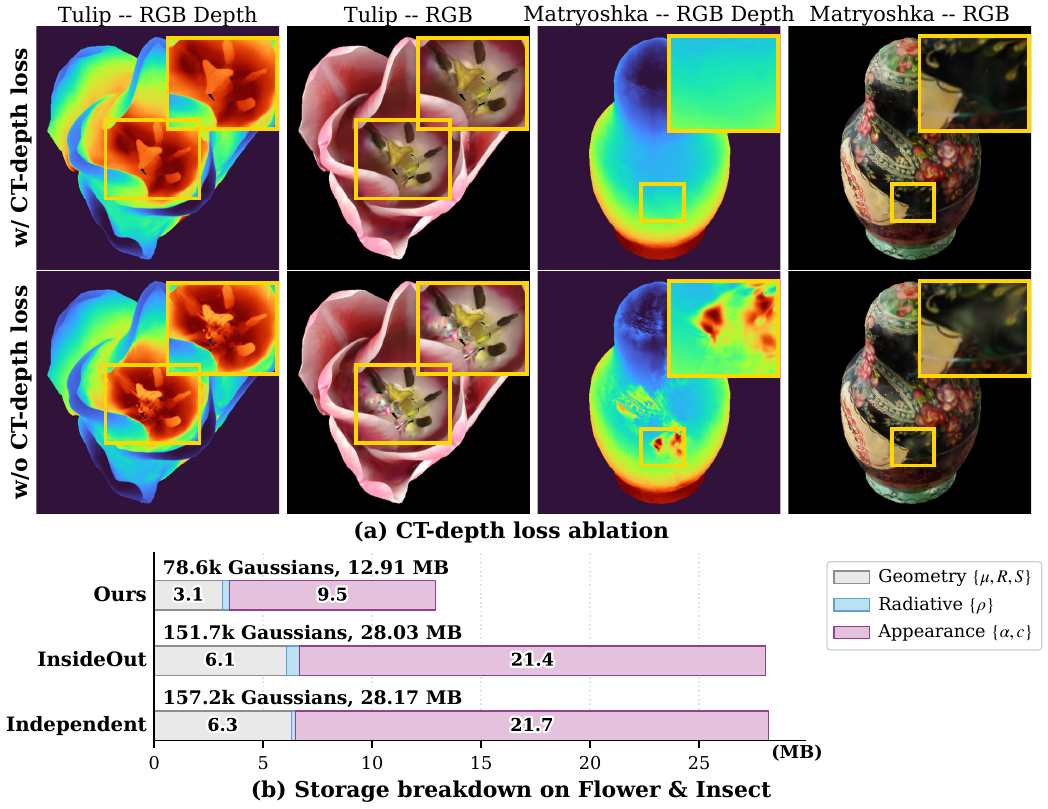}
\caption{\textbf{(a) CT-depth loss ablation:} removing the loss introduces visible depth artifacts.
\textbf{(b) Storage breakdown on Flower \& Insect:} Ours uses ${\sim}2.2\times$ less storage than Independent.
}
\label{fig:ablation_storage}
\end{figure}

\textit{CT-depth supervision (a $\to$ b).}
Disabling the CT-depth consistency loss reduces average RGB PSNR by around 0.16 dB. The loss anchors RGB surface to the X-ray-derived geometry, preventing surface primitives from drifting toward photometric optima that violate the geometric anchor; the resulting artifacts are visible in Fig.~\ref{fig:ablation_storage}(a).

\textit{Surface-set mechanism (b $\to$ c).}
Disabling the surface-set mechanism preserves rendering quality but inflates storage by roughly $2.5\times$. Without the surface set, appearance attributes (spherical harmonics) are allocated to every primitive, including interior primitives that contribute nothing to RGB rendering of opaque objects, so these parameters are wasted.

\textit{Geometry-based alignment (c $\to$ d).}
We replace our geometry-based alignment with the InsideOut-style hierarchical alignment~\cite{lee2025insideout} on optimized Gaussians, then run our joint optimization on top of the resulting transformation. RGB quality drops by roughly 0.9~dB PSNR while X-ray quality remains stable, and the resulting surface misalignment artifacts are visible in the middle column of Fig.~\ref{fig:qual_overlay}. Our purely geometric alignment is therefore more robust, and benefits downstream joint optimization.

\textit{Unified representation (a vs.\ e).}
Compared with training fully separate single-modality models (row (e)), our unified representation achieves higher RGB rendering quality without compromising X-ray reconstruction. By sharing geometry, both modalities reinforce the same structure, benefiting RGB in particular, where photometry alone offers only indirect geometric signal. This gain is not recovered by simply freezing the CT reconstruction and using it as a depth prior for a separate RGB set ($+0.06$\,dB over 3DGS; see the supplementary material). Combined with the storage savings noted above (with a larger ${\sim}2.2\times$ gap on Flower \& Insect; Fig.~\ref{fig:ablation_storage}(b)), the unified design outperforms the Independent baseline on every measured metric. Finally, X-ray quality remains stable across rows (a)--(d), confirming that staged optimization decouples X-ray fidelity from RGB supervision.

\textit{Robustness.}
We further stress-test the framework under noisy, sparse, and limited-angle X-ray acquisition, alignment derived from degraded CT geometry, and sparse RGB supervision. The results in Sec.~C of the supplementary material show that the framework remains robust across all these regimes. X-ray fidelity stays on par with the X-ray-only baseline, alignment and RGB quality are essentially unaffected, and the RGB gain from the shared CT geometry widens as views become sparser.

\section{Conclusion}

We presented a unified Gaussian framework for joint reconstruction of external surface appearance and internal volumetric structure. A single set of Gaussian primitives shares geometric parameters across both modalities, with appearance attributes restricted to surface-proximal primitives; cross-modal registration uses clean geometric proxies; and joint optimization is complemented by an X-ray-derived surface-depth consistency loss. Our design improves RGB rendering quality, yields more robust cross-modal alignment than post-hoc strategies, preserves X-ray fidelity, and is substantially more compact than two independent models.

\textit{Limitations and future work.}
First, our appearance model assumes that surface reflection dominates the RGB signal and therefore restricts appearance attributes to surface-proximal primitives. This assumption does not hold for transparent objects such as glass, or for materials with strong subsurface scattering or transmission (e.g., biological tissue or wax), whose appearance arises from light interacting with the object interior. Supporting such materials would require a volumetric appearance model, and paired RGB--X-ray datasets containing them are not yet available; we leave this to future work. Second, our framework depends on a reasonably accurate X-ray reconstruction to anchor the shared geometry, and may degrade when the X-ray inputs are extremely noisy or blurry. Third, current joint training improves RGB but matches the single-modality baseline on X-ray; extending RGB supervision to also benefit X-ray reconstruction is left as future work.

\begin{acks}
This work was supported by JST CRONOS JPMJCS25K6, JST-Mirai Program JPMJMI23G1, JSPS Grant-in-Aid JP23K16921, UTokyo-Google AI Symbiotic Future Society Program, Japan, Royal Society ISPF International Collaboration Award (ICA\textbackslash R2\textbackslash 252130), and Advanced AI Talent Development to Lead the Next-Generation AI for Intelligent Society (BOOST NAIS) of The University of Tokyo. We thank Kenji Kohiyama for contributing the insect data.
\end{acks}

\clearpage
\bibliographystyle{ACM-Reference-Format}
\bibliography{sample-base}

\ifwithsupp
\appendix
\renewcommand{\thesection}{\Alph{section}}
\renewcommand\thesubsection{\thesection.\arabic{subsection}}
\clearpage

\section*{Supplementary Materials}
\section*{Table of Contents}
\startcontents
\printcontents{ }{1}{}
\setcounter{section}{0}
\setcounter{equation}{0}

\section{Dataset Details}
\label{supp:dataset}

This appendix provides per-scene specifications and additional acquisition and rendering details that complement the Datasets paragraph in the main paper.

\subsection{Per-scene Specifications}
\label{supp:per_scene}

Table~\ref{tab:supp_dataset_summary} reports per-scene RGB train and test view counts under our $1{:}3$ split, the RGB resolution used at training and evaluation time, and the CT volume sizes used for X-ray synthesis. All scenes share the same X-ray sampling protocol ($50$ training and $120$ testing projections of $512\times 512$ pixels, uniformly sampled over $[0^\circ,
360^\circ)$).

\begin{table}[h]
\centering
\small
\caption{Per-scene dataset specifications. RGB Views (Tr/Te) lists train and test counts under our $1{:}3$ split. RGB Resolution is the resolution used at training and evaluation (raw images are downsampled to a width of $800$ pixels). CT Volume Size is the voxel grid at which we synthesize X-ray projections; for InsideOut it is the released volume downsampled to a uniform $256^3$, while for Flower and Insect we keep the native voxel grid.}
\label{tab:supp_dataset_summary}
\resizebox{1.0\linewidth}{!}{%
\begin{tabular}{l|l|c|c|c}
\toprule
Source & Scene & RGB Views (Tr / Te) & RGB Resolution & CT Volume Size \\
\midrule
\multirow{4}{*}{InsideOut~\cite{lee2025insideout}}
  & Matryoshka & 39 / 14 & $800\times 800$ & $256\times 256\times 256$ \\
  & Pharaoh    & 49 / 17 & $800\times 800$ & $256\times 256\times 256$ \\
  & Skull      & 36 / 13 & $800\times 800$ & $256\times 256\times 256$ \\
  & Toy Gun    & 75 / 26 & $800\times 800$ & $256\times 256\times 256$ \\
\midrule
\multirow{2}{*}{Flower~\cite{ijiri2014flower}}
  & Eustoma    & 60 / 20 & $800\times 800$ & $310\times 211\times 230$ \\
  & Tulip      & 60 / 20 & $800\times 800$ & $487\times 321\times 337$ \\
\midrule
\multirow{3}{*}{Insect}
  & E02        & 27 /  9 & $800\times 600$ & $427\times 704\times 797$ \\
  & E15        & 27 /  9 & $800\times 600$ & $455\times 671\times 870$ \\
  & E32        & 27 /  9 & $800\times 600$ & $453\times 668\times 868$ \\
\bottomrule
\end{tabular}%
}
\end{table}

\subsection{X-ray Synthesis Geometry}
\label{supp:xray_sim}

We use TIGRE~\cite{biguri2016tigre} with cone-beam geometry to synthesize all X-ray projections, with the geometric parameters listed in Table~\ref{tab:supp_xray_params}. Each ground-truth CT volume is normalized so that its longest physical extent equals $2$ (i.e., lies in $[-1,+1]$), with the remaining axes proportionally scaled to preserve the original aspect ratio; InsideOut volumes are cubic ($256^3$) and therefore fill the unit cube exactly, while Flower and Insect volumes occupy a proportionally smaller region along their shorter axes. The Insect scenes use a slightly larger detector ($3.5\times3.5$ vs.~$3.0\times3.0$ scene units) because their elongated bodies require a wider field of view. We use a clean (noise-free) synthesis setting to remain faithful to the released ground-truth volumes. Robustness under noisy, sparser, and limited-angle projections is evaluated in Sec.~\ref{supp:robustness}.

\begin{table}[h]
\centering
\small
\caption{Cone-beam X-ray geometry used in TIGRE. Linear quantities are given in normalized scene units (the longest extent of each object equals $2$).}
\label{tab:supp_xray_params}
\begin{tabular}{l|c|c}
\toprule
Parameter & InsideOut / Flower & Insect \\
\midrule
Source-to-origin (DSO)        & $5.0$  & $5.0$  \\
Source-to-detector (DSD)      & $7.0$  & $7.0$  \\
Detector resolution           & $512\times 512$ & $512\times 512$ \\
Detector physical extent      & $3.0\times 3.0$ & $3.5\times 3.5$ \\
Angular range                 & $[0^\circ, 360^\circ)$ & $[0^\circ, 360^\circ)$ \\
\#Training / Testing views    & 50 / 120 & 50 / 120 \\
\bottomrule
\end{tabular}
\end{table}

\subsection{RGB Capture and Rendering}
\label{supp:rgb_capture}

\paragraph{InsideOut.} We use the original RGB photographs released by~\cite{lee2025insideout}. Native resolutions vary by scene (from $1830\times 1830$ to $2597\times 2597$); for both training and evaluation, we downsample all views to $800\times 800$ pixels.

\paragraph{Flower.} We render RGB views with Blender ($800\times 800$, $512$ samples per pixel) from the \textsc{.blend} assets distributed with~\cite{ijiri2014flower}. Camera positions follow a smooth top-to-bottom spiral covering $5$ revolutions across an elevation band of $[-30^\circ, 85^\circ]$, yielding $80$ views per scene.

\paragraph{Insect.} We capture $36$ RGB photographs of each prepared insect specimen at native resolution $2048\times 1536$ and downsample to $800\times 600$ for training and evaluation, preserving the $4{:}3$ aspect ratio.

\paragraph{Foreground masks.} Foreground masks are required by our RGB rendering loss and metric computation. For the Flower scenes, we directly use the alpha channel produced by Blender. For InsideOut and Insect, we annotate masks with X-AnyLabeling~\cite{xanylabeling}, an open-source labeling tool that integrates the Segment Anything Model~\cite{kirillov2023segment,ravi2024sam}; the SAM-predicted masks are then manually refined to remove spurious holes and over-segmented backgrounds. PSNR / SSIM / LPIPS for RGB are computed only inside the foreground mask.

\paragraph{Initial point cloud.} As initialization for our unified Gaussian model, we run an MVS reconstruction (\textsc{MVSFormer}~\cite{cao2022mvsformer}) on the training RGB views of each scene rather than the COLMAP sparse cloud used by 3DGS / InsideOut. The resulting dense point cloud is filtered against the per-view foreground masks (keeping only points seen inside the mask in at least one view), uniformly downsampled by Furthest Point Sampling (FPS), and used both as the alignment source (paired with the X-ray-derived surface) and as the color initialization for the RGB Gaussian splats.

\subsection{Flower \& Insect Dataset Gallery}
\label{supp:gallery}

Figure~\ref{fig:supp_dataset_gallery} previews the five Flower \& Insect scenes used in our experiments. For each scene, we show two representative RGB photographs and two X-ray projections from different angles. The two Flower scenes (Tulip, Eustoma) are rendered from the \textsc{.blend} assets of~\cite{ijiri2014flower}; the three Insect scenes (E02, E15, E32) are captured in our lab from prepared specimens (see Sec.~\ref{supp:rgb_capture}).

\begin{figure}[h]
\centering
\includegraphics[width=\linewidth]{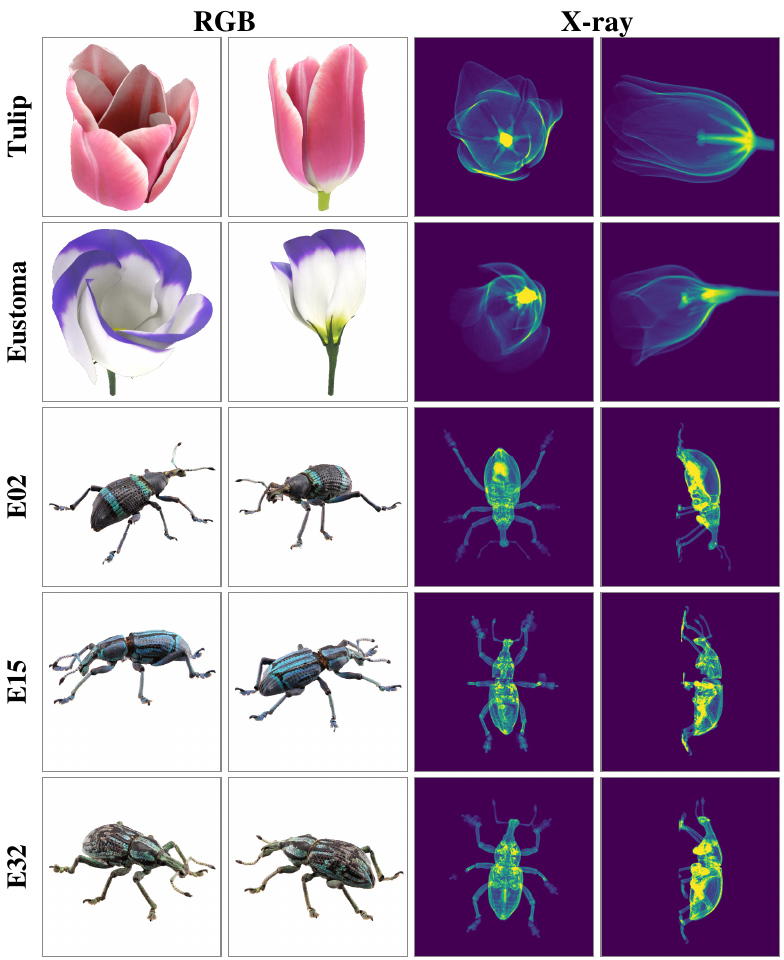}
\caption{\textbf{Flower \& Insect dataset gallery.} Each row shows one scene with two RGB photographs (left) and two X-ray projections (right) at different angles. X-ray projections are visualized with the viridis colormap.}
\label{fig:supp_dataset_gallery}
\end{figure}

\section{Implementation Details}
\label{supp:impl}

This appendix supplements the Implementation Details paragraph in the main paper with the full optimization procedure, complete optimizer settings, densification and pruning thresholds, and the remaining hyperparameters that the main paper does not list explicitly. We reuse the symbols defined in Sec.~\ref{sec:method}.

\subsection{Full Optimization Procedure}
\label{supp:algorithm}

For ease of reproduction, Algorithm~\ref{alg:full-pipeline} consolidates the procedural flow described in Sec.~\ref{sec:method} into a single self-contained pseudocode listing.
The algorithm is organized into the same four logical blocks as the main paper: a coarse X-ray pretraining (Stage~1, Sec.~\ref{subsec:joint-optimization}), the cross-modal geometry-based alignment (Sec.~\ref{subsec:alignment}), the surface-aware initialization that bridges the two stages (Sec.~\ref{subsec:joint-optimization}), and the joint optimization loop with periodic CT-mesh refresh (Stage~2, Secs.~\ref{subsec:joint-optimization}--\ref{subsec:optimization-strategy}).
We reuse the symbols defined in Sec.~\ref{sec:method} and refer to the loss equations there rather than restating them.
Concrete iteration budgets are $N_1{=}3{,}000$ and $N_2{=}12{,}000$, with mesh-refresh interval $T_{\mathrm{mesh}}{=}500$ and virtual-camera count $|\mathcal{C}^{\mathrm{geo}}|{=}180$; the remaining hyperparameters ($\tau_{d}$, $\sigma$, the loss weights, learning rates, and densification/pruning thresholds) are reported in the subsections below.

\begin{algorithm}[h]
\caption{Unified RGB-X-ray Gaussian Optimization}
\label{alg:full-pipeline}
\small
\begin{algorithmic}[1]
\Require RGB views $\{I^{\mathrm{rgb}}_{k}, \pi^{\mathrm{rgb}}_{k}\}_{k=1}^{K_{C}}$; X-ray projections $\{I^{\mathrm{xray}}_{l}\}_{l=1}^{K_{X}}$; FDK initialization $\mathcal{G}^{\mathrm{init}}$; iteration budgets $N_{1}, N_{2}$; mesh-refresh interval $T_{\mathrm{mesh}}$; surface threshold $\tau_{d}$.
\Ensure Unified Gaussian set $\mathcal{G} = \{\mathcal{G}_{i}\}_{i=1}^{N}$ with shared geometry $\{\mu_{i}, R_{i}, S_{i}\}$, density $\rho_{i}$, and surface-restricted appearance $\{\alpha_{i}, c_{i}\}_{i \in \mathcal{S}}$.
\vspace{2pt}
\State \textbf{// Stage 1: X-ray Pretraining} \Comment{Sec.~\ref{subsec:joint-optimization}}
\State $\mathcal{G} \gets \mathcal{G}^{\mathrm{init}}$; instantiate only $\{\mu_{i}, R_{i}, S_{i}, \rho_{i}\}$
\For{$\mathrm{iter} = 1$ \textbf{to} $N_{1}$}
    \State Sample X-ray view $c_{x}$ from $\{I^{\mathrm{xray}}_{l}\}$
    \State Render $\hat{I}^{\mathrm{xray}}(c_{x})$ from $\mathcal{G}$
    \State Compute $\mathcal{L}_{\mathrm{stage1}}$ \Comment{Eq.~\eqref{eq:stage1-loss}}
    \State Backward; X-ray-driven densify \& prune \Comment{R$^{2}$-Gaussian protocol~\cite{zha2024r}}
\EndFor
\State $\mathcal{G}^{\mathrm{coarse}} \gets \mathcal{G}$
\vspace{2pt}
\State \textbf{// Geometry-based Alignment} \Comment{Sec.~\ref{subsec:alignment}}
\State $V \gets \mathrm{Voxelize}(\mathcal{G}^{\mathrm{coarse}})$; binarize, morphologically close, retain largest connected component
\State $\mathcal{M}^{\mathrm{ct}} \gets \mathrm{MarchingCubes}(V)$; $\mathcal{P}^{\mathrm{ct}} \gets \mathrm{Sample}(\mathcal{M}^{\mathrm{ct}})$ \Comment{X-ray surface proxy}
\State $\mathcal{P}^{\mathrm{mvs}} \gets \mathrm{MVSFormer}(\{I^{\mathrm{rgb}}_{k}, \pi^{\mathrm{rgb}}_{k}\})$ \Comment{RGB surface proxy}
\State $T^{\mathrm{pca}} \gets \mathrm{PCAAlign}(\mathcal{P}^{\mathrm{mvs}}, \mathcal{P}^{\mathrm{ct}})$
\State $T \gets \mathrm{ICP}_{\mathrm{sim3}}(\mathcal{P}^{\mathrm{mvs}}, \mathcal{P}^{\mathrm{ct}};\ T^{\mathrm{pca}})$
\For{$k = 1$ \textbf{to} $K_{C}$}
    \State $E^{\mathrm{rgb}\to\mathrm{ct}}_{k} \gets E^{\mathrm{rgb}}_{k} T^{-1}$ \Comment{Eq.~\eqref{eq:camera-transform}}
\EndFor
\vspace{2pt}
\State \textbf{// Surface-aware Initialization} \Comment{Sec.~\ref{subsec:joint-optimization}}
\State Compute $d_{i} \gets \mathrm{dist}(\mu_{i}, \mathcal{M}^{\mathrm{ct}})$ for all $i$
\State $\mathcal{S} \gets \{\,i : d_{i} < \tau_{d}\,\}$ \Comment{Eq.~\eqref{eq:surface-set}}
\For{$i \in \mathcal{S}$}
    \State $\alpha_{i} \gets \alpha_{\mathrm{surf}}\cdot\exp(-d_{i}^{2}/2\sigma^{2})$ \Comment{Eq.~\eqref{eq:opacity-init}}
    \State $c_{i} \gets \mathrm{color}(\mathrm{NN}_{\mathcal{P}^{\mathrm{mvs}}}(\mu_{i}))$
\EndFor
\State $\mathcal{C}^{\mathrm{geo}} \gets \mathrm{FibonacciSphere}(n_{\mathrm{vir}};\ r{=}\mathrm{DSO},\ |\mathrm{elev}|{\le}75^{\circ})$
\vspace{2pt}
\State \textbf{// Stage 2: Joint Optimization} \Comment{Secs.~\ref{subsec:joint-optimization}--\ref{subsec:optimization-strategy}}
\For{$\mathrm{iter} = 1$ \textbf{to} $N_{2}$}
    \If{$\mathrm{iter} \bmod T_{\mathrm{mesh}} = 1$}
        \State $\mathcal{M}^{\mathrm{ct}} \gets \mathrm{MarchingCubes}(\mathrm{Voxelize}(\mathcal{G}))$ \Comment{periodic refresh}
    \EndIf
    \State Sample X-ray view $c_{x}$, training RGB view $c_{r}$, virtual view $c_{g}\sim\mathcal{C}^{\mathrm{geo}}$
    \State Render $\hat{I}^{\mathrm{xray}}(c_{x})$, $\hat{I}^{\mathrm{rgb}}(c_{r})$, depths $D^{\mathrm{rgb}}(c_{g})$ \Comment{swap $c_i{\to}z_i$ in Eq.~\eqref{eq:rgb-rendering}, restricted to $\mathcal{S}$}
    \State Rasterize $D^{\mathrm{ct}}(c_{g})$ from $\mathcal{M}^{\mathrm{ct}}$
    \State Compute $\mathcal{L}^{\mathrm{xray}}_{\mathrm{photo}}(c_{x})$, $\mathcal{L}^{\mathrm{rgb}}_{\mathrm{photo}}(c_{r})$, $\mathcal{L}_{\mathrm{TV}}$, $\mathcal{L}_{\mathrm{geo}}(c_{g})$ \Comment{Eq.~\eqref{eq:geo-loss}}
    \State Compute $\mathcal{L}_{\mathrm{stage2}}$ \Comment{Eq.~\eqref{eq:stage2-loss}}
    \State Backward
    \State Identify densification set $\mathcal{D}$ \Comment{Eq.~\eqref{eq:densify-set}; RGB branch restricted to $\mathcal{S}$}
    \State Identify pruning set $\mathcal{R}$ \Comment{Eq.~\eqref{eq:prune-set}; both attributes low}
    \State Densify, prune, and demote $\mathcal{S}$-members with $\alpha_{i}{<}\alpha_{\min}$
    \State Optimizer step
\EndFor
\State \Return $\mathcal{G}$
\end{algorithmic}
\end{algorithm}

A few aspects of Algorithm~\ref{alg:full-pipeline} merit emphasis. (i)~The X-ray photometric loss is retained throughout Stage~2 rather than frozen, so that RGB gradients cannot drift the shared geometry at the cost of X-ray fidelity. (ii)~The geometric loss $\mathcal{L}_{\mathrm{geo}}$ is evaluated on a virtual viewpoint $c_{g}$ sampled from a Fibonacci sphere $\mathcal{C}^{\mathrm{geo}}$, decoupled from the training RGB view $c_{r}$ used for the photometric loss; this exposes the surface to a denser angular cover than the training cameras alone. (iii)~The CT mesh $\mathcal{M}^{\mathrm{ct}}$ is re-extracted every $T_{\mathrm{mesh}}$ iterations from the current $\mathcal{G}$, so $D^{\mathrm{ct}}$ tracks the evolving shared geometry rather than freezing at the Stage~1 estimate. (iv)~Densification is mode-aware: X-ray gradients can spawn primitives anywhere in the volume, while RGB gradients can only do so within the surface set $\mathcal{S}$, preserving the surface/interior separation introduced in Eq.~\eqref{eq:unified-primitive}.

\subsection{Optimizer and Learning Rates}
\label{supp:lrs}
We optimize all parameters with Adam~\cite{kingma2014adam} ($\beta_1{=}0.9$, $\beta_2{=}0.999$). The view-dependent color $c_i$ is parametrized by spherical harmonics up to degree $3$ (the 3DGS default). Per-attribute learning rates follow 3DGS / R\textsuperscript{2}-Gaussian defaults (Table~\ref{tab:supp_lrs}).

\begin{table}[h]
\centering
\small
\caption{Per-attribute learning rates. Position, density, scaling, and rotation rates are linearly decayed from their initial to final values over $30{,}000$ steps.}
\label{tab:supp_lrs}
\begin{tabular}{l|c}
\toprule
Parameter & Learning rate (init $\to$ final) \\
\midrule
Position $\mu$        & $2{\times}10^{-4} \to 2{\times}10^{-5}$ \\
Density $\rho$        & $1{\times}10^{-2} \to 1{\times}10^{-3}$ \\
Scaling $S$           & $5{\times}10^{-3} \to 5{\times}10^{-4}$ \\
Rotation $R$          & $1{\times}10^{-3} \to 1{\times}10^{-4}$ \\
SH features $c$       & $2.5{\times}10^{-3}$ (constant) \\
Opacity $\alpha$      & $5{\times}10^{-2}$ (constant) \\
\bottomrule
\end{tabular}
\end{table}

\subsection{Surface-aware Initialization}
\label{supp:surface_init}
The decay scale in Eq.~\eqref{eq:opacity-init} is set to $\sigma = 0.2 \cdot \tau_d = 0.004$, so that for $i \in \mathcal{S}$ the initial opacity falls from $\alpha_{\mathrm{surf}} = 0.5$ at $d_i{=}0$ to a band-edge value of $1{\times}10^{-3}$ at $d_i{=}\tau_d$. Primitives outside $\mathcal{S}$ receive a fixed interior opacity of $1{\times}10^{-5}$.

\subsection{Densification and Pruning}
\label{supp:densify_prune}
Densification follows the 3DGS schedule (interval $=100$ iterations, active from iter $500$ to iter $8{,}000$). The position-gradient thresholds in Eq.~\eqref{eq:densify-set} are $\theta^{\mathrm{xray}} = 5{\times}10^{-5}$ (R\textsuperscript{2}-Gaussian default) and $\theta^{\mathrm{rgb}} = 1{\times}10^{-6}$. Pruning thresholds in Eq.~\eqref{eq:prune-set} are $\alpha_{\min} = 5{\times}10^{-3}$ and $\rho_{\min} = 1{\times}10^{-5}$. Surface-set demotion uses the same opacity threshold $\alpha_{\min}$.

\subsection{Alignment Pipeline}
\label{supp:alignment_details}
The Marching Cubes step~\cite{lorensen1998marching} in Algorithm~\ref{alg:full-pipeline} uses an iso-level of $0.3$ on the Stage~1 density volume. The subsequent ICP refinement uses a voxel-down-sampled correspondence radius of $0.01$ and a maximum of $2{,}000$ iterations.

\subsection{Geometric Constraint}
\label{supp:geom_constraint}
Stage~2 mesh re-extraction reuses the same Marching Cubes settings as the alignment mesh in Sec.~\ref{supp:alignment_details}. The virtual-camera sphere radius of Algorithm~\ref{alg:full-pipeline} resolves to $5.0$ in scene units (matching the X-ray DSO; see Sec.~\ref{supp:xray_sim}), and the depth maps $D^{\mathrm{rgb}}$ and $D^{\mathrm{ct}}$ used in Eq.~\eqref{eq:geo-loss} are rendered at half the X-ray detector resolution.

\subsection{Hardware and Training Time}
\label{supp:hardware}
All experiments are run on a single NVIDIA RTX A6000 GPU ($48$~GB memory). For the InsideOut dataset, training time (Stage~1 X-ray pretraining + Stage~2 joint optimization, excluding evaluation) ranges from $14$ to $21$ minutes per scene, averaging $16$ minutes (Table~\ref{tab:supp_runtime}). The runtime correlates with the final number of Gaussian primitives, with Matryoshka taking the longest because of its layered nested-doll structure ($\sim$$215$k primitives at convergence) compared with $\sim$$50$--$150$k for the other three scenes. Flower and Insect scenes use larger native voxel grids than InsideOut's uniform $256^3$ volumes and therefore incur a proportionally higher cost in X-ray rasterization; total per-scene runtimes are within the same order of magnitude.

\begin{table}[h]
\centering
\small
\caption{Per-scene training time on the InsideOut dataset (single NVIDIA RTX A6000). Total includes Stage~1 ($3{,}000$ X-ray pretraining iterations) and Stage~2 ($12{,}000$ joint optimization iterations); evaluation time is excluded.}
\label{tab:supp_runtime}
\begin{tabular}{l|c}
\toprule
Scene       & Training time (min) \\
\midrule
Matryoshka  & $21.2$ \\
Pharaoh     & $14.3$ \\
Skull       & $14.7$ \\
Toy Gun     & $14.0$ \\
\midrule
Average     & $16.0$ \\
\bottomrule
\end{tabular}
\end{table}

\subsection{InsideOut Reproduction}
\label{supp:insideout_repro}
The official InsideOut implementation and its data preprocessing are not publicly available, so we reproduce the method from the descriptions and hyperparameters reported in its paper (Sec.~\ref{subsec:experimental_setup}). In addition, only part of its capture data is publicly released (e.g., $251$ of $555$ views for Matryoshka). The released captures span multiple physical placements of each object but carry no metadata relating the placements, so only the views from a single object placement can be registered into a consistent coordinate frame; the resulting per-scene view counts are listed in Table~\ref{tab:supp_dataset_summary}. Note that InsideOut reports RGB metrics only in its supplementary material, based on its full non-public captures. The discrepancy between those results and our reproduction in Table~\ref{tab:permod_insideout} therefore likely stems from the unreleased pipeline and the substantially smaller registrable view subset, and our reproduction may underestimate the method's original performance.

\section{Robustness Analysis}
\label{supp:robustness}

We stress-test the framework under degraded inputs: noisy, sparse, and limited-angle X-ray acquisition and alignment robustness on the InsideOut dataset, and sparse RGB supervision on Flower\,\&\,Insect. All settings follow the main-paper protocol unless noted.

\paragraph{Noisy X-ray projections.} We add noise to the training X-ray projections following R\textsuperscript{2}-Gaussian~\cite{zha2024r}, keeping the test projections clean (Table~\ref{tab:supp_rob_noise}). X-ray 3D-PSNR degrades by only ${\sim}0.5$\,dB and identically for Ours and R\textsuperscript{2}-Gaussian, while RGB changes by at most $0.04$\,dB per scene. The joint design therefore adds no fragility relative to the X-ray-only backbone, and X-ray noise does not leak into the RGB channel.

\begin{table}[h]
\centering
\small
\caption{Robustness to noisy X-ray projections on the InsideOut dataset.}
\label{tab:supp_rob_noise}
\resizebox{1.0\linewidth}{!}{%
\begin{tabular}{l|l|cccc|c}
\toprule
Metric & Method & Matryoshka & Pharaoh & Skull & Toy Gun & Avg. \\
\midrule
\multirow{4}{*}{X-ray 3D-PSNR$\uparrow$}
  & R\textsuperscript{2}-Gaussian (clean) & 33.35 & 42.28 & 37.09 & 40.28 & 38.25 \\
  & R\textsuperscript{2}-Gaussian (noisy) & 32.92 & 41.92 & 36.39 & 39.90 & 37.78 \\
  & Ours (clean) & 33.39 & 42.13 & 36.90 & 40.25 & 38.17 \\
  & Ours (noisy) & 32.94 & 41.68 & 36.30 & 39.78 & 37.68 \\
\midrule
\multirow{2}{*}{RGB PSNR$\uparrow$}
  & Ours (clean) & 30.37 & 30.89 & 32.42 & 36.68 & 32.59 \\
  & Ours (noisy) & 30.35 & 30.85 & 32.38 & 36.68 & 32.57 \\
\bottomrule
\end{tabular}%
}
\end{table}

\paragraph{Sparse X-ray projections.} We halve the training projections from the $50$ used in our standard protocol to $25$ (Table~\ref{tab:supp_rob_sparse}). Ours matches R\textsuperscript{2}-Gaussian across X-ray 2D-PSNR, 3D-PSNR, and surface CD, indicating that joint RGB supervision does not compromise X-ray under sparser acquisition.

\begin{table}[h]
\centering
\small
\caption{X-ray reconstruction with $25$ training projections (uniform over $360^\circ$) on the InsideOut dataset. Surface CD denotes the Chamfer distance to the ground-truth CT surface.}
\label{tab:supp_rob_sparse}
\resizebox{1.0\linewidth}{!}{%
\begin{tabular}{l|l|cccc|c}
\toprule
Metric & Method & Matryoshka & Pharaoh & Skull & Toy Gun & Avg. \\
\midrule
\multirow{2}{*}{X-ray 2D-PSNR$\uparrow$}
  & R\textsuperscript{2}-Gaussian & 42.57 & 51.20 & 44.68 & 45.67 & 46.03 \\
  & Ours & 42.66 & 51.05 & 44.59 & 45.65 & 45.99 \\
\midrule
\multirow{2}{*}{X-ray 3D-PSNR$\uparrow$}
  & R\textsuperscript{2}-Gaussian & 28.67 & 40.23 & 33.01 & 36.54 & 34.61 \\
  & Ours & 28.74 & 39.77 & 32.96 & 36.58 & 34.51 \\
\midrule
\multirow{2}{*}{Surface CD$\downarrow$}
  & R\textsuperscript{2}-Gaussian & 0.0035 & 0.0028 & 0.0046 & 0.0028 & 0.0034 \\
  & Ours & 0.0035 & 0.0029 & 0.0046 & 0.0028 & 0.0034 \\
\bottomrule
\end{tabular}%
}
\end{table}

\paragraph{Limited-angle X-ray acquisition.} We confine the training projections to a $120^\circ$ arc, so a $240^\circ$ wedge is never observed by X-ray while the RGB views still cover the full $360^\circ$ (Table~\ref{tab:supp_rob_limang}). Ours remains on par with R\textsuperscript{2}-Gaussian across all metrics, both in-arc and out-of-arc, with only marginal differences in either direction. The full-coverage RGB supervision thus neither completes the missing wedge nor measurably degrades the reconstruction, and the X-ray channel remains self-sufficient under limited-angle acquisition.

\begin{table}[h]
\centering
\small
\caption{Limited-angle X-ray reconstruction on the InsideOut dataset. Training projections are confined to a $120^\circ$ arc; X-ray 2D-PSNR is reported on test projections inside the arc (in-arc) and on the missing wedge (out-of-arc). The alignment is fixed to the clean full-view transformation.}
\label{tab:supp_rob_limang}
\resizebox{1.0\linewidth}{!}{%
\begin{tabular}{l|l|cccc|c}
\toprule
Metric & Method & Matryoshka & Pharaoh & Skull & Toy Gun & Avg. \\
\midrule
\multirow{2}{*}{X-ray 2D-PSNR$\uparrow$ (out-of-arc)}
  & R\textsuperscript{2}-Gaussian & 41.63 & 46.38 & 41.19 & 40.89 & 42.52 \\
  & Ours & 41.64 & 46.03 & 40.97 & 40.79 & 42.36 \\
\midrule
\multirow{2}{*}{X-ray 2D-PSNR$\uparrow$ (in-arc)}
  & R\textsuperscript{2}-Gaussian & 49.88 & 53.96 & 51.60 & 51.71 & 51.79 \\
  & Ours & 50.51 & 54.78 & 51.23 & 51.59 & 52.03 \\
\midrule
\multirow{2}{*}{X-ray 3D-PSNR$\uparrow$}
  & R\textsuperscript{2}-Gaussian & 27.45 & 36.62 & 31.63 & 34.88 & 32.64 \\
  & Ours & 27.35 & 36.18 & 31.52 & 34.88 & 32.48 \\
\midrule
\multirow{2}{*}{Surface CD$\downarrow$}
  & R\textsuperscript{2}-Gaussian & 0.0037 & 0.0030 & 0.0050 & 0.0035 & 0.0038 \\
  & Ours & 0.0038 & 0.0030 & 0.0049 & 0.0029 & 0.0036 \\
\bottomrule
\end{tabular}%
}
\end{table}

\paragraph{Alignment under degraded X-ray geometry.} The limited-angle setting above also affects the coarse X-ray reconstruction, and hence the CT mesh used for alignment. We therefore re-derive the full PCA+ICP alignment from this limited-angle mesh instead of the clean full-view one (Table~\ref{tab:supp_rob_align}). RGB changes by at most $0.07$\,dB per scene, with X-ray 2D-PSNR and surface CD likewise unaffected, showing that the geometric-proxy alignment is not fragile to degraded X-ray geometry.

\begin{table}[h]
\centering
\small
\caption{Alignment robustness under limited-angle X-ray on the InsideOut dataset. Both settings use our full model, with the alignment fixed to the clean full-view transformation (fx) or re-derived by PCA+ICP from the CT mesh of the $120^\circ$ limited-angle reconstruction (al).}
\label{tab:supp_rob_align}
\resizebox{1.0\linewidth}{!}{%
\begin{tabular}{l|l|cccc|c}
\toprule
Metric & Alignment & Matryoshka & Pharaoh & Skull & Toy Gun & Avg. \\
\midrule
\multirow{2}{*}{RGB PSNR$\uparrow$}
  & fx (clean $T$) & 30.36 & 30.83 & 32.42 & 36.66 & 32.57 \\
  & al (derived $T$) & 30.43 & 30.84 & 32.37 & 36.63 & 32.57 \\
\midrule
\multirow{2}{*}{X-ray 2D-PSNR$\uparrow$ (out-of-arc)}
  & fx (clean $T$) & 41.64 & 46.03 & 40.97 & 40.79 & 42.36 \\
  & al (derived $T$) & 41.59 & 46.00 & 41.20 & 40.88 & 42.42 \\
\midrule
\multirow{2}{*}{Surface CD$\downarrow$}
  & fx (clean $T$) & 0.0038 & 0.0030 & 0.0049 & 0.0029 & 0.0036 \\
  & al (derived $T$) & 0.0038 & 0.0030 & 0.0050 & 0.0035 & 0.0038 \\
\bottomrule
\end{tabular}%
}
\end{table}

\paragraph{Sparse RGB views.} We reduce the RGB training views of each Flower\,\&\,Insect scene from the original $27$--$60$ (Table~\ref{tab:supp_dataset_summary}) to $15$ (Table~\ref{tab:supp_rob_fewview}). Our gain over single-modality 3DGS widens from $+2.72$\,dB at full views to $+4.16$\,dB, peaking on the thinnest structures (Tulip $+5.49$, Eustoma $+6.65$\,dB). These results indicate that the benefit of the X-ray modality to RGB appearance reconstruction becomes more pronounced under sparse RGB supervision, with the shared CT geometry helping most exactly where purely visual reconstruction is weakest.

\begin{table}[h]
\centering
\small
\caption{RGB reconstruction with $15$ training views on the Flower\,\&\,Insect dataset (X-ray unchanged). The 3DGS baseline uses FDK initialization, the same initialization source as Ours; the full-view gap is $+2.72$\,dB (Table~\ref{tab:permod_flower_insect}).}
\label{tab:supp_rob_fewview}
\resizebox{1.0\linewidth}{!}{%
\begin{tabular}{l|l|ccccc|c}
\toprule
Metric & Method & Tulip & Eustoma & E02 & E15 & E32 & Avg. \\
\midrule
\multirow{3}{*}{PSNR$\uparrow$}
  & 3DGS (15 views) & 28.10 & 26.81 & 29.73 & 29.13 & 30.18 & 28.79 \\
  & Ours (15 views) & 33.59 & 33.46 & 32.86 & 32.16 & 32.69 & 32.95 \\
  & $\Delta$ (Ours $-$ 3DGS) & $+5.49$ & $+6.65$ & $+3.13$ & $+3.03$ & $+2.51$ & $+4.16$ \\
\midrule
\multirow{2}{*}{SSIM$\uparrow$}
  & 3DGS (15 views) & 0.9596 & 0.9611 & 0.9644 & 0.9647 & 0.9691 & 0.9638 \\
  & Ours (15 views) & 0.9807 & 0.9806 & 0.9773 & 0.9775 & 0.9780 & 0.9788 \\
\midrule
\multirow{2}{*}{LPIPS$\downarrow$}
  & 3DGS (15 views) & 0.0662 & 0.0735 & 0.0323 & 0.0287 & 0.0279 & 0.0457 \\
  & Ours (15 views) & 0.0379 & 0.0500 & 0.0284 & 0.0268 & 0.0287 & 0.0344 \\
\bottomrule
\end{tabular}%
}
\end{table}

\FloatBarrier

\begin{figure*}[h]
\centering
\includegraphics[width=\textwidth]{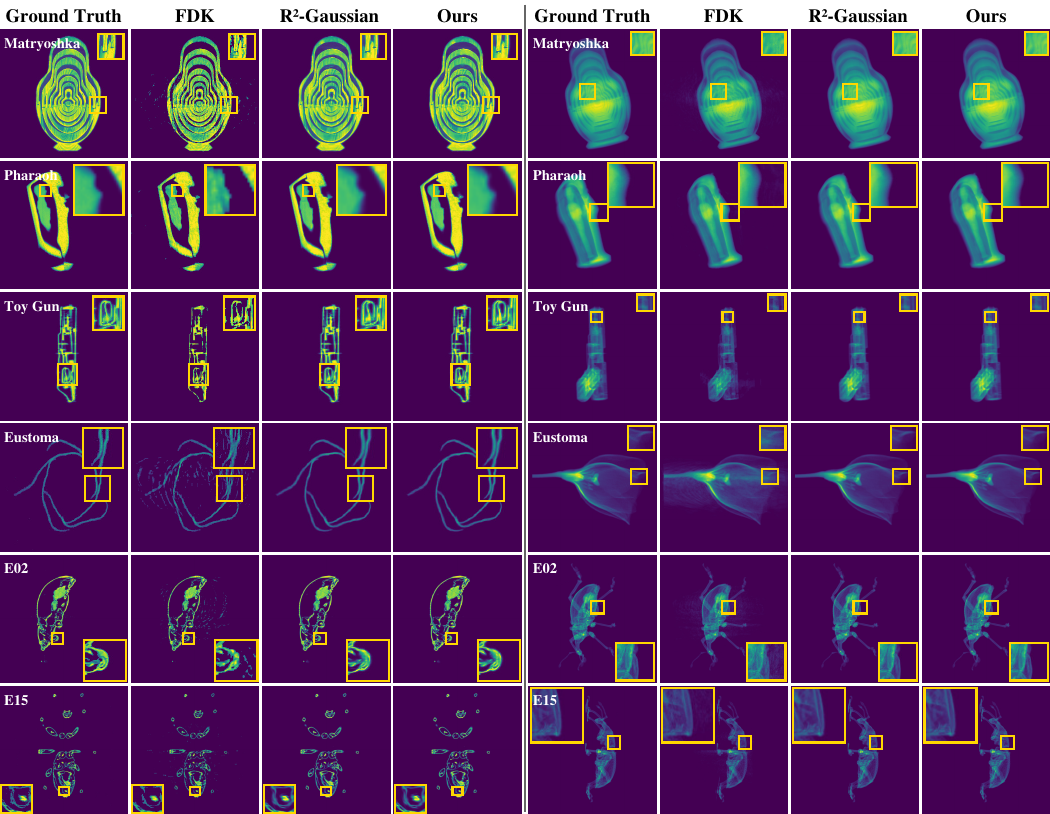}
\caption{\textbf{Additional CT slice and X-ray projection results (supplemental to Fig.~\ref{fig:qual_ct_xray}).}
Six additional scenes with the same eight-column layout. Findings match the main figure: our unified model reproduces R$^2$-Gaussian's fidelity on both modalities, while FDK shows the expected closed-form noise floor.}
\label{fig:qual_ct_xray_supp}
\end{figure*}

\section{More Results}

This appendix presents an additional quantitative comparison and qualitative results on additional scenes that complement the four main-paper qualitative figures: Fig.~\ref{fig:qual_ct_xray_supp} for CT slice and X-ray projection, Fig.~\ref{fig:qual_rgb_supp} for RGB novel-view synthesis, Fig.~\ref{fig:qual_overlay_supp} for unified RGB-X-ray overlay, and Fig.~\ref{fig:qual_alignment_supp} for cross-modal alignment.

\paragraph{Joint optimization vs.\ a frozen CT prior.} We evaluate the alternative of reconstructing CT once, freezing it, and using it only as a depth prior for a completely separate RGB Gaussian set, keeping the same frozen Stage-1 CT mesh, alignment, and initialization as Ours. As shown in Table~\ref{tab:supp_frozen_prior}, this variant is essentially identical to the single-modality 3DGS baseline ($+0.06$\,dB) and $1.33$\,dB below our joint model. A frozen CT prior therefore recovers almost none of the joint gain, which stems from jointly refining the shared geometry under both modalities' supervision.

\begin{table}[h]
\centering
\small
\caption{RGB PSNR of a separate RGB Gaussian set with a frozen CT depth prior on the InsideOut dataset. The 3DGS and Ours rows are from Table~\ref{tab:permod_insideout}.}
\label{tab:supp_frozen_prior}
\resizebox{1.0\linewidth}{!}{%
\begin{tabular}{l|cccc|c}
\toprule
Method & Matryoshka & Pharaoh & Skull & Toy Gun & Avg. \\
\midrule
3DGS (MVS init) & 29.56 & 29.64 & 31.08 & 34.54 & 31.20 \\
Separate RGB set + frozen CT prior & 29.77 & 29.63 & 31.06 & 34.59 & 31.26 \\
Ours (joint) & 30.37 & 30.89 & 32.42 & 36.68 & 32.59 \\
\bottomrule
\end{tabular}%
}
\end{table}

\begin{figure*}[h]
\centering
\includegraphics[width=\textwidth]{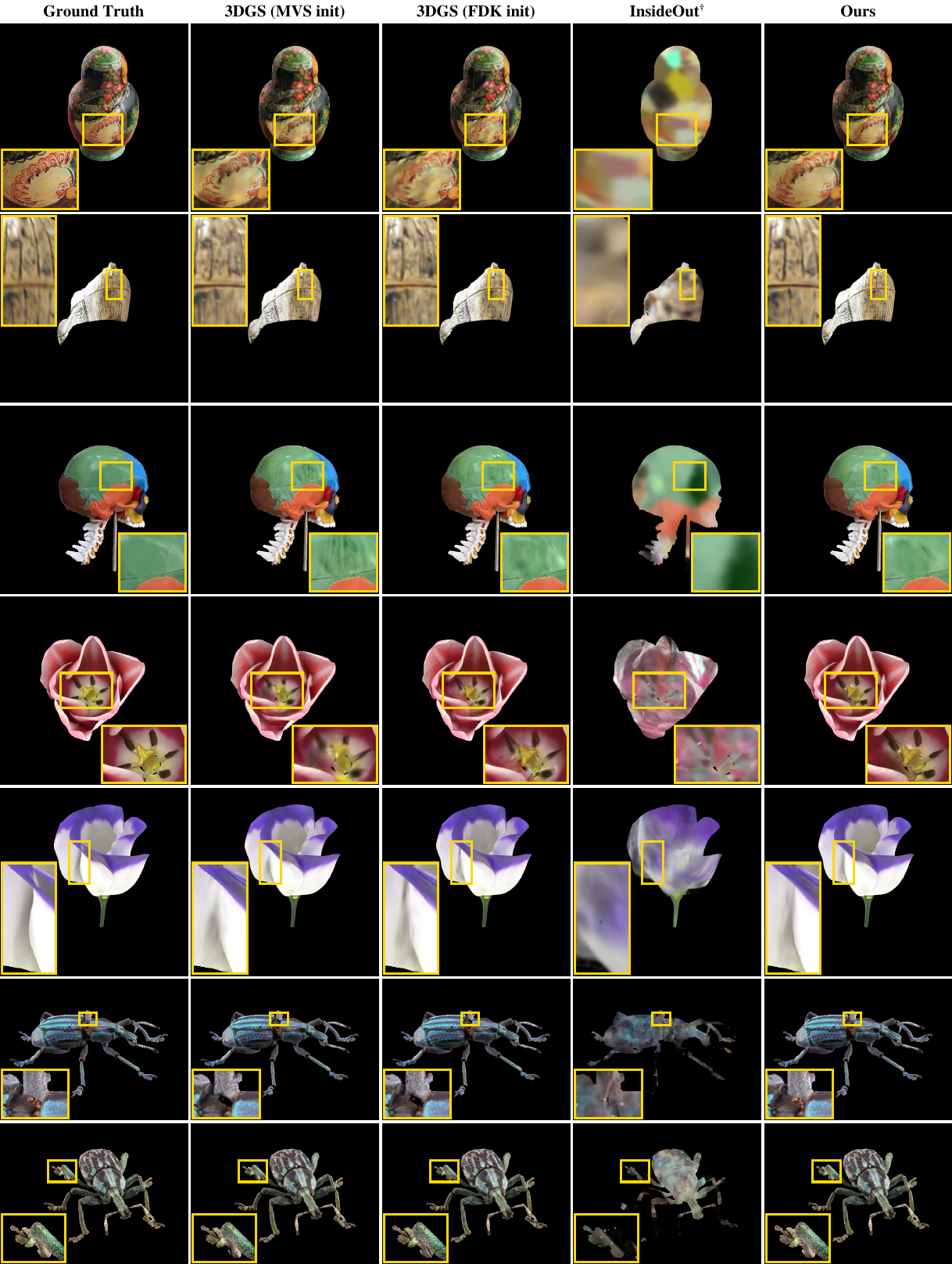}
\caption{\textbf{Additional RGB novel-view synthesis scenes (supplemental to Fig.~\ref{fig:qual_rgb}).}}
\label{fig:qual_rgb_supp}
\end{figure*}

\begin{figure*}[h]
\centering
\includegraphics[width=\textwidth]{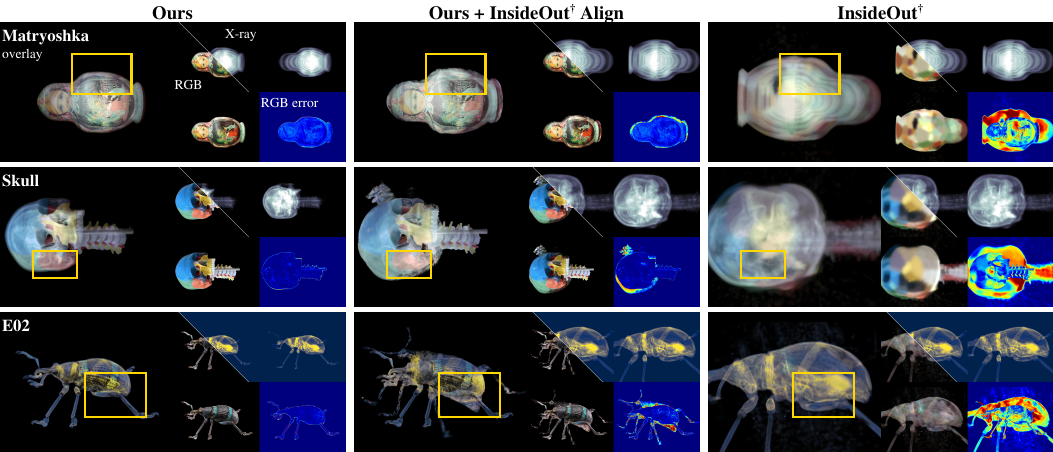}
\caption{\textbf{Additional unified RGB-X-ray showcase scenes (supplemental to Fig.~\ref{fig:qual_overlay}).}
Six additional scenes spanning both datasets, with the same three-panel cell layout. Findings match: Ours produces consistent inside/outside rendering across all scenes; Ours~+~IO-align retains the unified representation but with mild surface-alignment artifacts; our InsideOut reproduction shows consistently lower RGB quality, consistent with the main-paper results.}
\label{fig:qual_overlay_supp}
\end{figure*}

\begin{figure*}[h]
\begin{minipage}{0.48\textwidth}
\includegraphics[width=\linewidth]{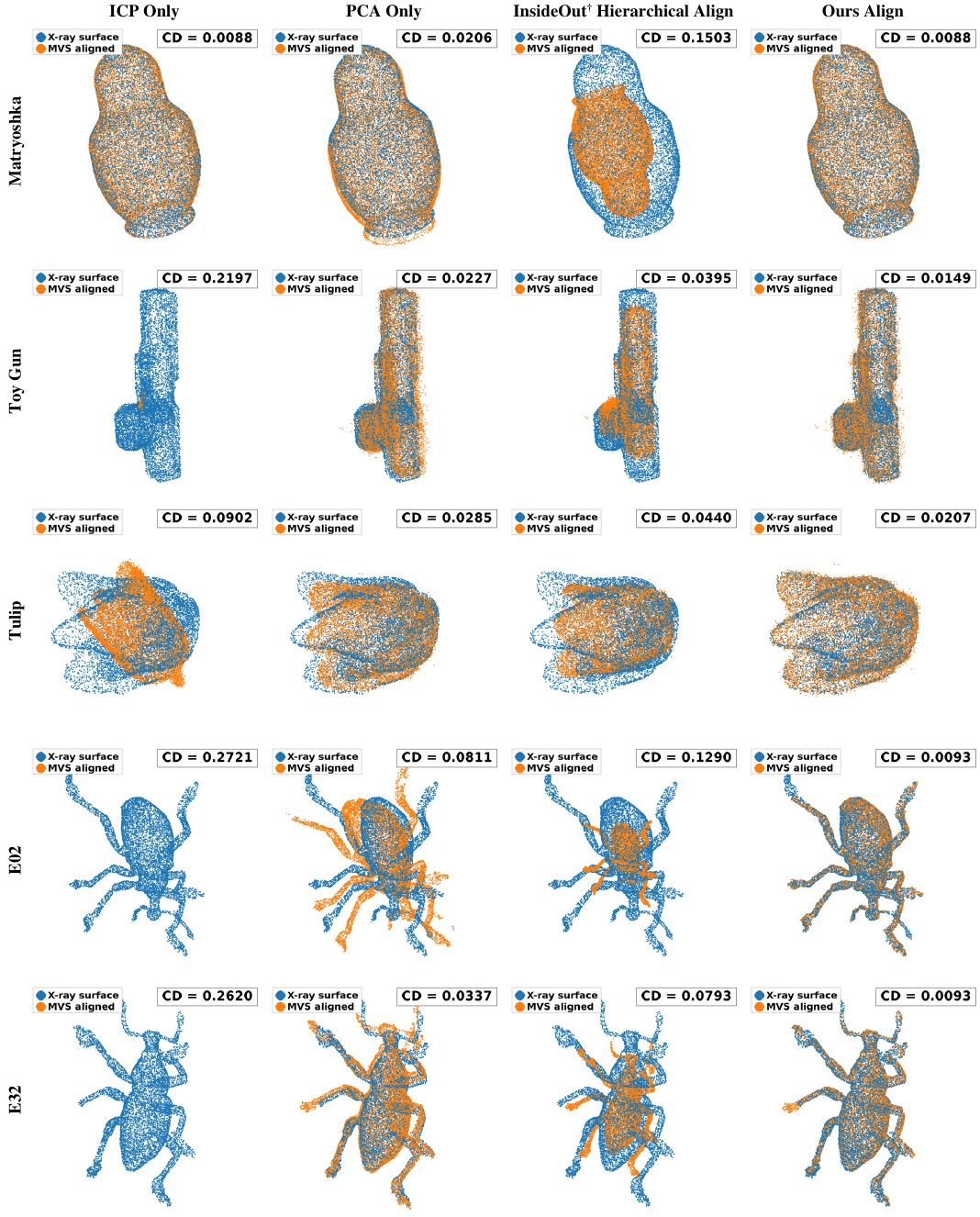}
\caption{
\textbf{Additional alignment-comparison scenes (supplemental to Fig.~\ref{fig:qual_alignment}).}}
\label{fig:qual_alignment_supp}
\end{minipage}
\end{figure*}

\fi

\end{document}